\documentclass[manuscript,screen,authorversion,nonacm]{acmart}

\usepackage{algorithm} %
\usepackage{algorithmic} %
\usepackage[shortlabels]{enumitem}
\usepackage{subcaption} %
\usepackage{multirow} %
\usepackage{float} %
\usepackage{xspace} %
\usepackage{xurl}
\usepackage{longtable}
\usepackage{wrapfig}
\usepackage{array}

\AtBeginDocument{%
  }

\setcopyright{acmlicensed}
\copyrightyear{2024}
\acmYear{2024}
\acmDOI{XXXXXXX.XXXXXXX}

\acmConference[Conference acronym 'XX]{Conference}{June 03--05,
  2024}{Woodstock, NY}
\acmISBN{978-1-4503-XXXX-X/18/06}

\begin{document}

\title[Generalized People Diversity]{Generalized People Diversity: Learning a Human Perception-Aligned Diversity Representation for People Images}

\author{Hansa Srinivasan}
\affiliation{%
  \institution{Google Research}
  \country{United States}
}

\author{Candice Schumann}
\authornote{Both authors contributed equally to this research.}
\affiliation{%
  \institution{Google Research}
  \country{United States}
}

\author{Aradhana Sinha}
\authornotemark[1]
\affiliation{%
  \institution{Google Research}
  \country{United States}
}

\author{David Madras}
\affiliation{%
  \institution{Google Research}
  \country{United States}
}

\author{Gbolahan Oluwafemi Olanubi}
\affiliation{%
  \institution{Google Research}
  \country{United States}
}

\author{Alex Beutel}
\affiliation{%
  \institution{Open AI}
  \country{United States}
}

\author{Susanna Ricco}
\affiliation{%
  \institution{Google Research}
  \country{United States}
}

\author{Jilin Chen}
\affiliation{%
  \institution{Google Research}
  \country{United States}
}

\newcommand{\todo}[1]{ {\color{red}{TODO}: #1}}
\newcommand{\cs}[1]{ {\color{red}{Candice}: #1}}
\newcommand{\hs}[1]{ {\color{blue}{Hansa}: #1}}
\newcommand{\as}[1]{ {\color{orange}{Anu}: #1}}
\newcommand{\dm}[1]{ {\color{olive}{David}: #1}}
\newcommand{\sr}[1]{ {\color{purple}{ricco}: #1}}

\newcommand{\eg}{\textit{e}.\textit{g}.}
\newcommand{\dct}[1]{ {\color{darkgray}\textbf{#1}\vspace{7pt}}}
\newcommand{\link}[2]{{\color{blue}\href{#1}{#2}}}

\newcommand{\sign}{\text{sgn}}
\newcommand{\ourmethod}[0]{Perception-Aligned Text-derived Human representation Space }
\newcommand{\Ourmethod}[0]{Perception-Aligned Text-derived Human representation Space }
\newcommand{\ourmethodabbv}[0]{PATHS }
\newcommand{\ourdataset}[0]{Diverse People Dataset }
\newcommand{\ourdatasetabbv}[0]{DPD }

\begin{abstract}

Capturing the diversity of people in images is challenging: recent literature tends to focus on diversifying one or two attributes, requiring expensive attribute labels or building classifiers.
We introduce a diverse people image ranking method which more flexibly aligns with human notions of people diversity in a less prescriptive, label-free manner.
The \ourmethod (PATHS) aims to capture all or many relevant features of people-related diversity, and, when used as the representation space in the standard Maximal Marginal Relevance (MMR) ranking algorithm \cite{carbonell1998use}, is better able to surface a range of types of people-related diversity (e.g. disability, cultural attire).
\ourmethodabbv is created in two stages. First, a text-guided approach is used to extract a person-diversity representation from a pre-trained image-text model. 
Then this representation is fine-tuned on perception judgments from human annotators so that it captures the aspects of people-related similarity that humans find most salient. 
Empirical results show that the \ourmethodabbv method achieves diversity better than baseline methods, according to side-by-side ratings from human annotators.

\end{abstract}




\maketitle

\section{Introduction}\label{sec:introduction}

Image ranking systems often mediate how we access and discover images --- through web-based recommendation \citep{wang2012multimodal,jing2008visualrank,yu2014learning}, dataset curation \citep{deng2009imagenet,yang2020towards}, and even image generation systems \citep{yu2022scaling}.
When we query these systems for images of people, what the system surfaces can shape our perceptions by reinforcing or dismantling stereotypes~\cite{kay2015unequal}.
As such, there is growing interest to make the top-ranked results for people-seeking queries more diverse. ~\cite{silva2023representation,karako2018using,celis2020implicit,noble-2018,otterbacher2018investigating}.

\begin{wrapfigure}{R}{0.4\textwidth}
    \centering
    \includegraphics[width=0.4\textwidth]{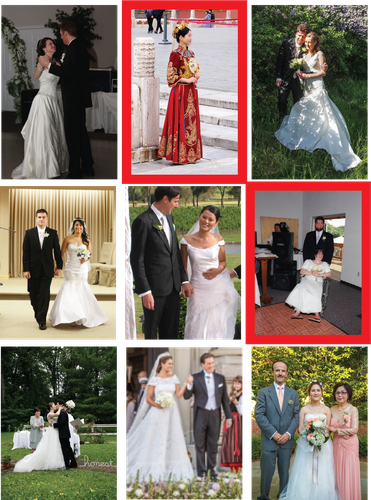}
    \caption{An example of image ranking results for the query of ``Bride'' with our proposed \ourmethodabbv method. This method promotes the images outlined in red: a bride in culturally Chinese attire, and a bride in a wheelchair.
    }
    \label{fig:qualitative_paths_only}
    \vspace{-20pt}
\end{wrapfigure}

The term ``diversity'' carries two distinct meanings in the context of image ranking. Firstly, in traditional recommendation systems, which employ image ranking, ``diversity'' means providing users with a range of visually novel, but still relevant, results to their query~\cite{van2009visual,carbonell1998use,celis2016fair,qian2017image}. 
Secondly, from a sociological perspective, ``diversity'' refers to people~\cite{plaut2015new,ashkanasy2002diversity}. 
While exact definitions vary, ``diversity'' in this paper is used to refer to the variety of sociocultural identities (e.g., cultural backgrounds, lifestyles, nationalities, religions) represented within a group of people.  The goal of the image recommendation systems in this paper is to achieve diversity in the second sense while maintaining quality: to surface query-relevant images that also showcase a range of perceived sociocultural identities. Though this paper is grounded in an image recommendation use case, this work may also be of interest in other areas that require evaluating diversity in a set of images such as dataset curation.

\begin{wrapfigure}{L}{0.6\textwidth}
    \centering
    \includegraphics[width=0.6\textwidth]{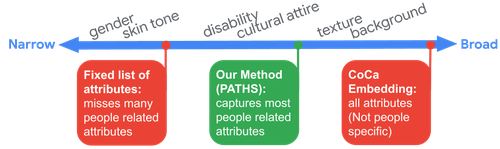}
    \caption{A spectrum of people image diversification methods: from narrow (considers too \textit{few} attributes of people diversity) to broad (considers too \textit{many} attributes of visual diversity).
    At the top of this figure, we list where various attributes tend to fall on this spectrum, with attributes such as gender presentation being most prevalent in narrow settings due to their more frequent availability.
    }
    \label{fig:spectrum_of_diversification}
    \vspace{-10pt}
\end{wrapfigure}

Improving person-diversity in an image ranking or recommendation system presents a key design decision: how should one encode sociocultural identity information about people in a way that the ranking algorithm can understand and effectively use? 
In Fig \ref{fig:spectrum_of_diversification} demonstrates a spectrum of options, highlighting two contrasting failure modes.
One naive approach is to treat \textit{all} visual information equally and apply standard diversification techniques (e.g., submodular maximization, MMR \cite{borodin2012max,carbonell1998use}). This, however, leads to overly \textit{broad} diversification where the algorithm might focus on elements like image backgrounds or styles, instead of prioritizing the diversity of the people in the images~\cite{celis2016fair}.

Recent work has addresses this by narrowly focusing on a few structured "sensitive attributes" for diversification \cite{karako2018using,silva2023representation, TanjimGansForImageSearchDiversity}. Often, this means diversifying based on perceived gender expression and/or skin tone. While this is better than the naive approach, it has a key limitation: defining diversity through a few specific attributes restricts how richly we can represent the diversity of people. Consequently, existing work in this area often reduces diversity \textit{narrowly} to only gender expression and/or skin tone \cite{celis2020implicit, TanjimGansForImageSearchDiversity}. This simplification is at odds with the complex way we socially perceive diversity, which is influenced by a far wider range of sociocultural factors~\cite{jackson2003recent} like ethnicity~\cite{rattan2013diversity}, perceived gender expression~\cite{lindqvist2021gender}, cultural attire~\cite{reeves2012muslim,ghumman2013religious}, body shape~\cite{de2018body}, age~\cite{kunze2011age} and more.

This "narrow" approach to diversification, which focuses on a limited set of pre-defined "sensitive attributes," has three algorithmic limitations. Firstly, it is inherently a prescriptive process that depends on the algorithm designers' own understanding of what diversity ``matters''. This can introduce biases that exclude non-normative images (e.g. women without long hair, people in wheelchairs) \cite{denton2021genealogy,prabhu2020large}. Secondly, the reliance on a fixed list of attributes (usually with discrete labels and equal weighting among attributes) makes the system inflexible and unable to fully capture the complexity of human diversity. Finally, the requirement to label every image for each attribute~\cite{schumann2023consensus} is a costly, ethically challenging, and difficult task to scale with legal and privacy risks~\cite{andrews2023view,prabhu2020large,wu2020gender}. The availability of existing attribute data during the design stage further biases the system towards specific well-studied attributes. Collectively these limitations can result in biased outcomes, inflexibility, and ethical challenges that limit creating a truly inclusive representation of diversity.

Given the limitations of overly "narrow" (fixed attribute list) and excessively "broad" (all visual diversity) diversification, the goal is to find a balanced way to represent sociocultural identity.
This paper introduces \textit{\ourmethod} (\ourmethodabbv), a continuous representation space for people images.
Unlike narrow fixed-attribute-list methods, it avoids the need for predefined attributes, or prescriptive discrete labels or similarity metrics, reducing the potential for bias from the system designer, and the need for costly labeling.
Importantly, similarity in \ourmethodabbv aligns with human perception of similarity between people in images. This means demographically similar people in different settings (ex: Gottfried Leibniz writing at a desk and Isaac Newton sitting under an apple tree) should be close together, while demographically dissimilar individuals in similar settings (ex: Gottfried Leibniz writing at a desk and Katherine Johnson writing at a desk) would be further apart in the representation space.

\ourmethodabbv is learned as a refinement of a general image-text embedding \cite{yu2022coca}, with two fine-tuning steps that prioritize flexibility in the types of diversity we can learn without attribute labels: \textit{text-guided subspace extraction}, where we learn a low-dimensional projection intended to remove non-person information, and \textit{perception alignment}, where this projected subspace is further modified to align more closely with human-annotated diversity judgments.
Experiments demonstrate how \ourmethodabbv achieves the right balance of diversification (See Fig. \ref{fig:spectrum_of_diversification}) through showing strong performance on two contrasting datasets: the canonical Occupations dataset\cite{celis2020implicit} which penalizes overly-broad diversification methods, and a new dataset: the Diverse People Dataset, which penalizes overly-narrow methods respectively.
\ourmethodabbv achieves the best average performance ($55.5\%$ diversity increase) across both datasets, demonstrating its effectiveness as a representation space for comprehensive people-diversity.
Moreover, qualitative case studies demonstrate instances where \ourmethodabbv better captures the specific diversity types human annotators find important.

\subsection{Contributions}
This paper includes the following contributions:
\begin{itemize}[leftmargin=*,noitemsep]
    \item \textbf{A method to learn a representation space that can be used for diverse ranking of people images. Notably, this method does not require or encourage predefining a notion of diversity through a short list of strictly-taxonomized attributes (e.g. perceived gender expression, skin tone), or training on photos labeled for such attributes.} 
    This method consists of a standard diversification algorithm applied to our new \textit{\ourmethod} (PATHS).
    \item Empirical results showing that  \ourmethodabbv achieves the strongest ranking diversity across both the Occupations dataset\cite{celis2020implicit}, and the new Diverse People Dataset created for this paper, showing both the ability to diversify along a range of people attributes, as well as to discount non-people attributes in diversification, achieving a good tradeoff between narrow and broad diversification methods.
    \item Evidence that perceptions of diversity are complex: human annotators give more weight to some people-related attributes over others in diversity perception judgements. Case studies show where the \ourmethodabbv method---unlike other methods---agrees with human perception by correctly weighting these people-related attributes more.
    
\end{itemize}

\section{Related Work}\label{sec:related_work}
\paragraph{Diversity in Ranking and Recommendations.}
    
Significant work exists on producing diverse rankings or subsets in web systems. These works seek to balance diversity with relevance using similarity metrics. Examples include DPPs, MMR, and submodular maximization models~\cite{carbonell1998use,celis2018fair,borodin2012max}. However, as noted by \citet{celis2018fair}, general visual similarity metrics (e.g. ImageNet pre-trained image embeddings) fail to generate diversity along attributes that contribute to people diversity, such as gender expression or skin tone.

To bridge this gap, several papers focus on diversifying image sets along attributes of people diversity that are socially salient (e.g. gender, skin tone, etc.).
\citet{karako2018using} diversifies gender expression only by using a small set of gender expression-labeled images. 
\citet{silva2023representation} uses attribute-specific classifiers --- with a focus on skin tone --- to diversify a large-scale recommendation system. 
\citet{celis2020implicit}, on the other hand, move away from labeled attributes and attribute classifiers by using diverse hand-curated ``control sets.'' These control sets, however, do not generalize to new contexts and must be manually-designed for each new image set.
As such, they cannot scale to ranking web systems that must quickly serve new image sets. Finally, \citet{tanjim2022generating} address the problem of visual diversity by using GANs to generate image sets diverse in perceived gender and race, changing the problem setting from one of selecting diverse sets, to creating them from scratch.

Measuring the effect of these diversification schemes is also challenging, as the diversity of a group of people is inherently subjective: for instance, a person's perception that a group is diverse can vary depending how their identities relate to the ones present in the group \cite{danbold2020drawing}. Obtaining human annotations on subjective tasks is a large area of research~\cite{schumann2023consensus,goyal2022toxicity,aroyo2023reasonable,diaz2022crowd}. Much of this research shows the importance of diversity of the human annotators~\cite{schumann2023consensus,goyal2022toxicity,diaz2022crowd} and notes the effects of subconscious biases and human inconsistency across time~\cite{aroyo2023reasonable}. Following \citet{schumann2023consensus}, all human annotation tasks in this paper are completed by 8 annotators across 4 different geographical regions across the world.

\paragraph{Learning and Manipulating Embeddings.}
The core methodological component of this work is learning and manipulating a new representation space (i.e. an embedding), a topic touched on by several strains of related research.
One cornerstone of this work is leveraging pretrained image-text models such as CLIP \citep{radford2021learning} or CoCa \citep{yu2022coca}, where image and text features are aligned in a co-embedding space.
A number of recent works probe these pretrained models for the structure and compositionality of representations in their embedding spaces~\cite{lewis2022does,wolff2023independent,zhou2023clip}. For instance, \citet{wolff2023independent} find using PCA on text embeddings to identify representations for attributes like color, size, patterns, yield a corresponding visual sub-space for the concepts. \citet{zhou2023clip} are able to use PCA over a corpus of text embeddings to manipulate image embeddings along specific attributes of people. 

This work is also influenced heavily by lines of work in the fairness literature which consider manipulating embeddings through removing specified information.
The most similar works are linear concept removal \citep{bolukbasi2016man,ravfogel2022linear} which use a linear projection procedure to remove a single attribute (such as perceived gender expression) from a pretrained embedding.
\citet{chuang2023debiasing} also propose a projection-based approach, and further leverage an image-text co-embedding space to remove information about a single attribute.
A more involved set of approaches are presented in the fair representation learning literature, in which methods attempt to learn an embedding from scratch while removing the influence of a sensitive attribute, either through an adversary \citep{madras2018learning,edwards2015censoring} or regularization \citep{louizos2015variational,song2019learning}.
\ourmethodabbv also bears some similarities to approaches such as from \citet{creager2019flexibly}, who draws inspiration from the disentanglement literature to remove multiple attributes from an embedding simultaneously.

A small but growing literature considers using human annotations to learn representations directly.
Notably, \citet{andrews2023view} have the most similar goals to this paper --- they seek to produce a human-perception aligned embedding from human annotations of image triplets, while acknowledging the significant shortcomings in labeling human attributes or per-attribute classifiers for images. In order to make use of a limited amount of data, however, their images are restricted to cropped, frontal images of faces, which are a highly constrained domain and as such are less able to convey many attributes of diversity that are possible in the broad range of people images we are concerned with (full body, various poses and backgrounds), (e.g. cultural attire, body shape).
\citet{cui2016fine} also use human-in-the-loop annotations in metric learning to find a low-dimensional manifold (unrelated to diversity), sending high confidence examples for labelling and using them in the next round as either positives or hard negatives.

Finally, this work draws on the distance metric learning literature by using a triplet learning method to incorporate human perceptual feedback.
Triplet learning has been successful in both linear metric learning approaches \citep{chechik2010large,mei2014logdet}, and deep learning-based methods \citep{hoffer2015deep,wang2014learning}.
There are also similar pair-based approaches in the classical metric learning literature, where pairs which are labelled as similar/dissimilar are brought closer/further in embedding space \citep{hadsell2006dimensionality,chopra2005learning}.

\section{Proposed method}\label{sec:methods}
This paper's core contribution is a method for creating a representation space that can richly represent complex notions of people diversity, including attributes that are salient to real human annotators, without requiring numerous people attribute labels.
This space is called the \textit{\ourmethod (PATHS)}. Sections \ref{subsec:text-guided} and \ref{subsec:human-rated-triplets} describe these two stages involved in creating PATHS:
\begin{enumerate}[leftmargin=*,noitemsep]
    \item Text-guided subspace extraction: Starting with a pretrained image embedding, extract an embedding which contains only the person-diversity-relevant information.
    \item Perception alignment: Refine the embedding so that images which are closer together contain people who are considered more perceptually-similar according to human judgment.
\end{enumerate}
Finally, Section \ref{subsec:problem_setup} describes how \ourmethodabbv can be used to diversify an image ranking system (such as the one described in \citet{celis2020implicit}): we apply MMR \cite{carbonell1998use}, a standard diversification method, to select a set of images which are diverse within PATHS (additional evaluations of the space itself and any text datasets mentioned are provided in the Supplemental Materials).

\subsection{Text-guided approach to extracting a person diversity subspace} \label{subsec:text-guided}

\begin{wraptable}{r}{0.53\textwidth}
\centering
\setlength{\tabcolsep}{1pt} %
\begin{tabular}[width=\columnwidth]{lrr}
           & \multicolumn{2}{c}{Linear Probes(AUC)}  \\ 
Text-derived Subspace   & People & Non-people  \\ \toprule

Step 0: CoCa Embedding ($d=1408$)  &  $97\%$  &   $96\%$ \\ 
Step 1: People projection ($d=12$)  &  $92\%$  &   $83\%$ \\ 
Step 2: Project out background ($d=12$)  &  $91\%$  &   $80\%$ \\ 
\bottomrule
\end{tabular}
\caption{The goal is a representation of the original CoCa embedding that captures only attributes about people diversity. Such a representation should perform well on linear probes for ``People tasks,'' (those where the discriminative signal is expected to be people-related e.g. ``female doctor'' vs ``male doctor'') and not well on linear probes for ``Non-people tasks''
(e.g. ``person indoors'' vs ``person outdoors'').
Each step discards information from the original CoCa Embedding, reducing performance slightly on the ``person tasks''. 
However, performance on ``non-people tasks'' is reduced by a greater amount, suggesting
that more non-person information is discarded.
This table shows the dimensionality of the resulting embedding after each step.}

\label{tab:linear_probe_evals}
\vspace{-25pt}
\end{wraptable}

The cost of training a high-quality representation from scratch is prohibitive; we choose instead to start with CoCa text-image embeddings \cite{yu2022coca}, and perform two text-guided steps to extract a person-diversity representation, choosing CoCa as our starting point since it is trained on web-scale alt-text and annotated images and thus may already capture some notion of what humans notice about images. 
Recall that diversifying in the CoCa space directly would be too ``broad'' for diversification purposes; the first step in this method is therefore intended to narrow its scope to align more closely with person-diversity attributes.

\paragraph{Step 1: Project Onto People Attributes.} 
The high-level goal of this step is to find the subspace of the general image embedding that contains information which is relevant to people diversity.
Once this subspace is found, the general embedding can be projected onto that subspace, called a ``person-diversity representation''.
If the subspace learning process is successful, it will contain a more narrow set of information about the image than the full  embedding, but a broader set of human-related information than simply using perceived gender and skin tone labels.

To estimate this subspace without images or attribute labels, one can leverage CoCa's multimodal nature (where an image of an object and a text description of that object are frequently close in embedding space) and instead use a text set of phrases referring to people.
This text set is generated using Bard\footnote{https://bard.google.com/}\footnote{For more details see Appendix~\ref{app:datacard_bard}.} as a set of 100 nouns referring to people (e.g. ``bride'' or ``doctor'') and a set of adjectives applying to people across a range of diversity attributes (e.g. ``gender-fluid'' or ``Buddhist'').
For each noun, CoCa embeddings for all [adjective] + [noun] phrases are collected, and PCA was run on this set of embeddings, selecting the top $d_p$ components, where $d_p > 0$ is just the target size of the projected embedding.
Conceptualizing each individual noun's embeddings as a noisy observation of the ``person-diversity subspace'', the PCA components are averaged across all nouns to yield the projection matrix used to transform a CoCa embedding to a ``person-diversity representation''.

To collect the ``seed list'' of diversity attributes, we hosted a focus group with 18 experts in diversity, machine learning fairness, and responsible AI with the goal of understanding what kinds of visualizable diversity attributes are important in image retrieval. The experts were asked to enumerate diversity attributes that they thought were important. They were then split into smaller groups to discuss the visualizability and language surrounding those attributes. This focus group agreed to prioritize the following attributes of diversity: perceived gender expression, body type, disability, nationality, age, religion, and perceived sexual orientation. 
Note that while this method does perform an enumeration of attributes in this step, the design process is much more flexible than standard alternatives in the literature: it does not require a taxonomy or weighting scheme across these attributes, nor does it need expensive individually annotated images. It is low-cost to add a new attribute, thereby encouraging more expansive representations of diversity.

\paragraph{Step 2: Project Out Background.} 
In theory, the previous step could successfully project out all non-people-diversity information from the embedding.
However, in practice it is helpful to target a specific non-person attribute, background, in a second removal step.
Mimicking the previous approach, a set of locations (e.g. ``beach'', ``office'', ``outdoors'') were generated and systematically added these to the [adjective] + [noun] phrases (e.g ``bride at the beach''). Similar to above, PCA is applied on the CoCa text embeddings of these phrases. 
The top $d_b$ components were selected to form a ``background subspace'' which was further projected out from the person-diversity representation, choosing $0 \leq d_b < d_p$. 
The final result of these two projection steps was $P$, a linear transformation of the original CoCa image embedding (the product of two projection matrices) that projects to a ``text-derived person diversity subspace''.
See Appendix \ref{app:linear-probes} for a discussion of hyperparameter tuning procedures used for $d_b, d_p$.

\begin{wraptable}{R}{0.53\textwidth}
\vspace{-20pt}
\begin{tabular}{ll}
                          &                                                                                                                                           \\
\multirow{3}{*}{
\begin{minipage}{16mm}
   \includegraphics[width=16mm]{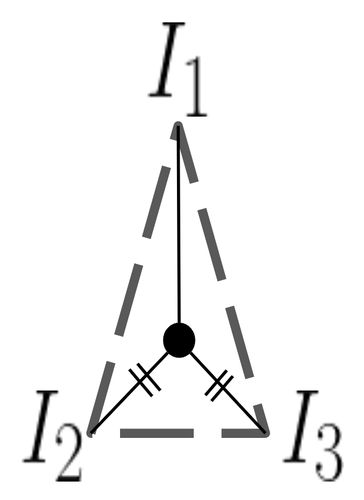}
\end{minipage}
} 
                          & 
                          \begin{tabular}[c]{@{}l@{}}
                          \textbf{Case 1:} $I_1$ is most different\\      
                          \hspace{4ex}$I_2$, $I_3$ are equally different
                          \end{tabular}                        \\
                          & 
                          \begin{tabular}[c]{@{}l@{}}
                          Human Annotator Most Different Votes:\\     
                          \hspace{4ex}$I_1: 4$,  $I_2:0$, $I_3: 0$\\     
                          \hspace{4ex}$I_1: 2$,  $I_2:1$, $I_3: 1$
                          \end{tabular}          \\
                          & 
                          \begin{tabular}[c]{@{}l@{}}
                          Relative Similarity Eqns:\\       
                          \hspace{4ex}$S(I_1, I_2) <  S(I_2, I_3)$,  $S(I_1, I_3) < S(I_2,I_3)$
                          \end{tabular}
                          \\
\midrule
\multirow{3}{*}{
\begin{minipage}{16mm}
   \includegraphics[width=16mm]{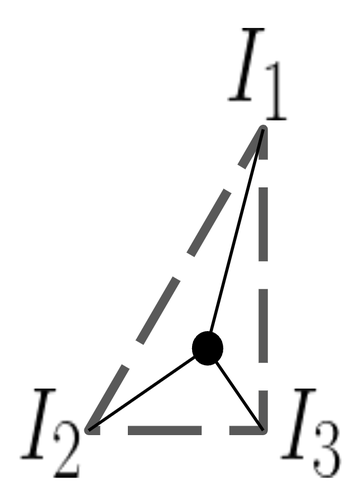}
\end{minipage}
} 
                          & 
                          \begin{tabular}[c]{@{}l@{}}
                          \textbf{Case 2:} $I_1$ is most different\\    
                          \hspace{4ex}$I_2$ is next most different
                          \end{tabular}                        \\
                          & 
                          \begin{tabular}[c]{@{}l@{}}
                          Human Annotator Most Different Votes:\\     
                          \hspace{4ex}$I_1: 3$,  $I_2:1$, $I_3: 0$\\     
                          \end{tabular}          \\
                          & 
                          \begin{tabular}[c]{@{}l@{}}
                          Relative Similarity Eqns:\\       
                          \hspace{4ex}$S(I_1, I_2) <  S(I_1, I_2) < S(I_1,I_3)$ \\
                          \end{tabular}
                          \\        
\midrule
\multirow{3}{*}{
\begin{minipage}{16mm}
   \includegraphics[width=16mm]{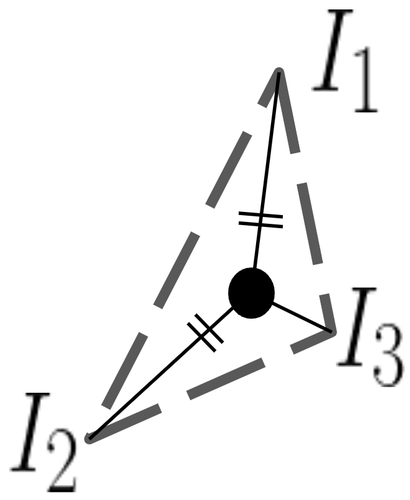}
\end{minipage}
} 
                          & 
                          \begin{tabular}[c]{@{}l@{}}
                          \textbf{Case 3:} $I_1$, $I_2$ are equally most different\\      
                          \end{tabular}                        \\
                          & 
                          \begin{tabular}[c]{@{}l@{}}
                          Human Annotator Most Different Votes:\\     
                          \hspace{4ex} $I_1: 2$,  $I_2:2$, $I_3: 0$\\     
                          \end{tabular}          \\
                          & 
                          \begin{tabular}[c]{@{}l@{}}
                          Relative Similarity Eqns:\\       
                          \hspace{4ex}$S(I_1, I_3) < S(I_1, I_2), S(I_2, I_3) < S(I_1, I_2)$\\     
                          \end{tabular}
                          \\    
\end{tabular}
    \caption{These figures describe the process for extracting a partial ordering on similarity from the annotation data, in which 4 human annotators pick which image of $I_1$, $I_2$, and $I_3$ is most different from the rest. 
    Here the degree of difference is visualized as the distance from the image to the centroid of a triangle created by the triplet in some space. This gives a relative ordering of the triangle edge lengths. Similarity between images $I_A$, $I_B$ is $S(I_A, I_B)$, which is inversely proportional to the length of the edge between the two images.}
    \label{tab:diff_to_sim}
    \vspace{-35pt}
\end{wraptable}

\subsection{Perception Alignment: Fine-tuned Representation using Human Perception}\label{subsec:human-rated-triplets}
The previous projection step is intended to isolate as much of the relevant information as possible in the extracted representation; however, there is no guarantee that distances in this embedding space (derived from CoCa) correspond to human perceptions of similarity of the people contained.
To this end, further fine-tuning was done by gathering human perceptions of people diversity\footnote{See Appendix~\ref{app:human_annotations} for a description of the human annotation task}, and using them to learn a linear transformation.

\subsubsection{Getting human perception annotations.} \label{subsubsec:get_human_triplet_ratings}
Each image triplet was annotated 4 times  (once in each of the 4 regions), with annotators selecting the image containing the person they considered to be the most different.
The annotators rated a dataset of 30k image tripets each, where 10k where sourced from the MIAP dataset \cite{schumann2021step}.
To increase a) the number of images used for training and b) the diversity of people represented in the images,  we source the next 20k image triplets from a separate scrape of Google Image Search with queries (not overlapping with the evaluation set) from the generated [adjective] + [noun] text phrases used to find the person diversity subspace.

\subsubsection{Metric learning: Fine-tuning with human annotations} \label{subsubsec:metric_learning}
The goal is to learn a %
matrix, $\hat{M}$, that is intended to project the people-diversity representation into a space that better aligns with human perception. 
$\hat{M}$ is trained with gradient descent on a modified triplet loss on the $n=30$k human annotations obtained in the previous section (85\%/10\%/5\% train-validation-test split). 
The final people-diversity representation --- PATHS --- for an image, $I$, is given by $e(I) = \text{CoCa\_Embedding}(I) \cdot P \cdot \hat{M}$. 

Human annotations are used as the ground truth for the relative similarities between three images in a triplet: each of four annotators identifies the image that is most different from the other two as described in Sec \ref{subsec:human-rated-triplets}.
Table \ref{tab:diff_to_sim} describes the procedure for converting these four human annotations into \textit{relative similarities} between images, letting $S(I_A, I_B) > 0$ denote the latent ``ground truth'' similarity between two images (Image $A$ and image $B$) based on the corresponding annotations.

Let the learned similarity function between two images be $1 - $the Euclidean distance in embedding space, $\hat{S}(I_A, I_B) = 1 - \|e(I_A)-e(I_B)\|$.
One can attempt to enforce the inferred ground truth inequality (where defined, as in Table \ref{tab:diff_to_sim}) between a pair of edges, $S(I_A, I_B)$ and $S(I_A, I_C)$ using a standard triplet loss with margin $\alpha$ anchored at $I_A$, the shared vertex~\cite{chechik2010large,mei2014logdet}:
\begin{equation}
\begin{aligned}
T(I_A, I_B, I_C) = \max\Bigl( & -1 \cdot \sign\bigl(S(I_A, I_B) - S(I_A, I_C)\bigl) \cdot \\
     &\bigl(\hat{S}(I_A, I_B) - \hat{S}(I_A, I_C) \bigl) + \beta, 0\Bigl )
\end{aligned}
\label{eqn:triplet_loss}
\end{equation}
The overall loss for an image triplet is then the sum of the three triplet losses, one anchored at each image:
\begin{equation}
\begin{aligned}
L(I_A, I_B, I_C, \mathbf{M}) = & T(I_A, I_B, I_C) + T(I_B, I_A, I_C) 
\\ &+ T(I_C, I_A, I_B) + \gamma \|\mathbf{M} - Id\|_1 + \lambda \|\mathbf{M} - Id\|_2^2
\end{aligned}
\end{equation}
Since $\hat{M}$ is being multiplied to produce the final embedding, the regularization happens against the identity matrix, $Id$.

\subsection{Diverse Ranking} \label{subsec:problem_setup}
\ourmethodabbv can be converted into a diverse ranking method by simply applying a diverse ranking algorithm to the learned \ourmethodabbv space.
We choose the simple Maximal Marginal Relevance (MMR) algorithm (Algorithm \ref{alg:mmr}) to replicate the set-up from Celis and Keswani's canonical work on the people-diversity task \cite{celis2020implicit}.

Modern large-scale image ranking systems typically follow a retrieval stage, which reduces the number of candidate images from a large corpus (of millions) to a much smaller set. In our experiments, we simulate this stage by using evaluation sets of ~100 images. These sets are described in section \ref{subsec:task_framework:tasks}.
After the retrieval stage, the 100 candidate images are ranked and MMR ~\cite{carbonell1998use} is used to select the top 9 to display in a 3x3 grid. 

\begin{algorithm}[H]
\caption{Maximal Marginal Relevance (MMR)}
\label{alg:mmr}
\begin{algorithmic}
\STATE Let $I$ denote an image. Let $S$ be the set of selected images, $AllImages$ refer to all images that may be selected, and $C$ refer to some in-distribution calibration set of images (we use $|C| = 10$k).
\STATE Let $k$ be the target number of diverse images to select (in experiments, we use $k = 9$).
\STATE Let Relevance$(I)$ be a black-box value of how relevant $I$ is to a given query (we implement this as in \citet{celis2020implicit}).
\STATE Let $e(I)$ be the learned person embedding (we use \ourmethodabbv).
\STATE Let $D$ be the Euclidean distance function.
\STATE Let $\mu_{De}$, $\sigma_{De}$ be the mean and standard deviation of $\{D(e(I_A),e(I_B)) \forall I_A, I_B \in C \times C\ \mid I_A \neq I_B \}$. \hspace{-3ex}
\STATE EmbedDistZScore$(e, I_A, I_B) = \frac{D(e(I_A),e(I_B)) - \mu_{De}}{\sigma_{De}}.$
\STATE MarginalDiversity$(I, S)=$
          $\textstyle\sum_{I_A \in S} \frac{\text{EmbedDistZScore}(e, I,I_A)}{|S|}$
\STATE MMR($I$, $S$) $= ( 1- \alpha)$ Relevance($I$)   + 
        \\\hspace{19.5ex} $\alpha$ MarginalDiversity($I$, $S$)
\STATE $S$ = \{ \}
\FORALL{$i \in 1, 2, \cdots,  k $:}
  \STATE $I_{new}=\textrm{argmax}_{I \in AllImages - S}\text{MMR}(I, S)$
  \STATE $S = S \cup \{I_{new}\}$
\ENDFOR
\end{algorithmic}
\end{algorithm}

The relevance score is the same as in ~\citet{celis2020implicit}, simply the average cosine similarity of the image to the top 10 ranked images from the dataset using a general image embedding \cite{simonyan2014very}. Marginal diversity is the normalized Euclidean distance between the new image and the centroid of the existing top images in some representation space. Normalization is done to ensure that different values of $\alpha$ have comparable effects for different representations of $e$. Note that while MMR is an existing method, the core contribution of this work is the method for creating the people-diversity representation used to compute marginal diversity, and these representations could be used with diversification methods other than MMR.

\section{Experimental Setup}

\subsection{Evaluation Datasets} \label{subsec:task_framework:tasks}

\begin{figure}
    \centering
    \includegraphics[width=6.5cm]{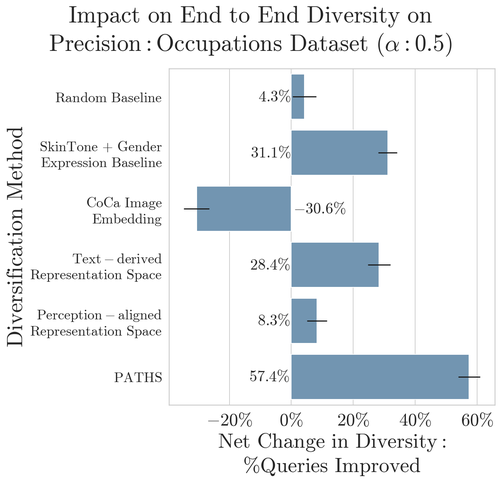}
    \hspace{2ex}
    \includegraphics[width=6.5cm]{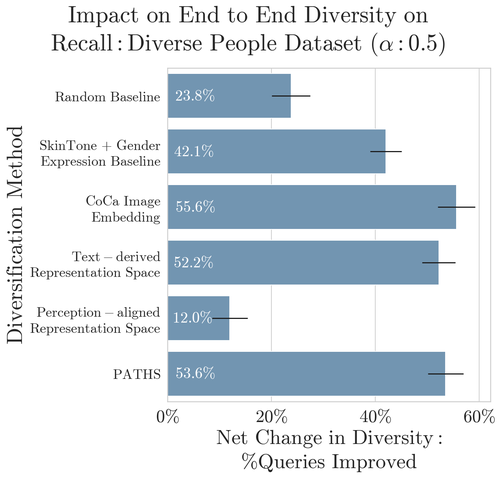}
    \caption{Across both datasets, \ourmethodabbv achieves the best added diversity ($\uparrow= $better).
    On the Occupations dataset, which penalizes overly-broad diversification, it outperforms all other methods.
    On the Diverse People dataset, which penalizes overly-narrow diversification, \ourmethodabbv, the text-derived space, and the base CoCa Embedding (the broadest diversification method) both outperform all other baselines.
    As a non-person specific embedding that encompasses all types of visual diversity, CoCa adds the most general visual diversity (hence high performance on Diverse People Dataset where all images are of people), but does not add people-specific diversity (hence poor performance on the Occupations dataset which contains many non-person images).
    }
    \vspace{-10pt}
    \label{tab:sxs}
\end{figure}

To reiterate from the introduction, the goal is to diversify image sets in a way which is neither too narrow or too broad; specifically, to diversify along as many axes of sociocultural diversity of people as possible, while \emph{not} to diversifying other aspects of non-people visual diversity.
The dual goals recommend a specific evaluation setup, which leverages two contrasting datasets.
The public Occupations dataset~\cite{celis2020implicit} is used to test diversification along \textit{only} features that people perceive as salient to people diversity, and not on non-people visual diversity. 
Additionally, we create a new dataset, the Diverse People Dataset (DPD), to test diversification along as many people-related attributes as possible.
As described below, the two qualitatively different datasets complement each other well --- overly-broad diversification methods should fail on Occupations, and overly-narrow diversification methods should fail on DPD.
Together, they enable the identification of a diversification method that strikes the balance between overly narrow and overly broad diversification -- capturing many different types of people-related diversity, without capturing irrelevant non-people related diversity. While image ranking is used in a variety of systems, we ground this evaluation in image recommendation: the datasets serve to simulate an image retrieval system and the task is to re-rank the images such that the top images are diverse.

\paragraph{Occupations Dataset:} Occupations is the canonical public baseline for the task of diversifying image sets of people. This dataset was created by scraping results from Google Image Search in 2020 for 96 occupations (such as custodian, paralegal, dentist, web developer) ~\cite{celis2020implicit}. 
Images in this dataset often do not have people as the main focus: people often occupy a very small part of the total image or are present as drawings/clip art. While these images' styles contribute to visual diversity, they don't necessarily add to person-diversity. Additionally, though the dataset does have a fair amount of perceived gender expression and skin tone diversity, the dataset under-represents people from non-western cultural backgrounds, non-slim body types, and people with disabilities. It can be viewed as the results of a retrieval stage with low availability of diversity, and used to evaluate how a ranking method would perform in this setting. We use this dataset to measure whether a diversification method promotes people-diversity over more broad (people-unrelated) types of visual diversity.
This task penalizes overly-\textit{broad} diversification: due to the high amount of visual variation in the images that is non-people-related, a general diversification method will likely not improve people diversity.
However, due to the small number of axes of people diversity represented (mostly perceived gender expression and skin tone), it is less able to assess whether a method has overly-\textit{narrow} diversification -- for this, the Diverse People Dataset is used.

\paragraph{\ourdataset (DPD):}
To evaluate how well the diversification method can surface many different kinds of people-diversity, we created the \ourdataset (DPD) by scraping 100 queries on Google Image Search in May 2023. These queries were carefully chosen to match real searches by people in the US\footnote{Note that the US-centric nature of this query set is a limitation --- see the Limitations section for more discussion.}, where the top 100 results were largely  close-up images of people (e.g., ``handsome man''). The 100 queries included references to roles (e.g "family", "bride"), professions (e.g "basketball player", "business person"), general terms for people (e.g. "woman", "toddlers), descriptors (e.g "funny person", "friendly person"), and fashion (e.g. "bangs hairstyle", "country club outfits").

Within each query, the top 80 images were scraped. Then a set of people attribute adjectives were added to each query (e.g., ``handsome plus sized man'' or ``handsome elderly man'') to scrape an additional $\sim10$ images with added people diversity. 
These attribute adjectives were chosen to cover a broad range of types of diversity, primarily covering the aspects of diversity identified as important by the focus group, while also yielding coherent high quality search results based on manual inspection. Note that this choice of sub-queries is one way to elicit sufficient diversity in the dataset, but there are many other viable choices as well. %
Additionally, a set of non-people attribute adjectives (e.g., ``handsome man in a suit'') were added to scrape $\sim10$ images that were more visually diverse, but irrelevant to people diversity --- this visual diversity is still significantly less than that in the Occupations dataset, where frequently people are not even the focus of the image.
The dataset of queries and their composition can be found in Appendix~\ref{app:datacard}. As this paper is grounded in the problem of "diverse ranking", and retrieval is outside of the scope of this paper, this dataset is used to comprehensively evaluate our ranking method by mocking a "high diversity retrieval system" in which sub-queries are used to help identify diverse candidates. 

As this dataset intentionally focuses on images where people are the primary subject of the photo, and where those people are diverse according to a wide range of attributes, it can evaluate a method's ability to diversify many people-related features -- that is whether the diversification method is \textit{broad} enough.
This dataset is less suitable to measure whether people-attributes are promoted over non-people attributes (for that we rely on the Occupations dataset).

\subsection{Baseline Methods}\label{subsec:task_framework:baselines}
\ourmethodabbv is tested against a number of baselines: each defining a representation space to be used in MMR when calculating Marginal Diversity.
\begin{itemize}
    \item \emph{Random Baseline}. A random baseline is used to demonstrate that the task is not trivially easy. Here MarginalDiversity is a random sample from a normal distribution. This baseline also serves as a sanity check for the human annotators annotating end to end diversity (Section \ref{sec:eval_rate_e2e}); annotators should rate higher for increased people diversity, not merely for any image difference from base.
    \item \emph{SkinTone + Perceived Gender Expression Baseline} Much of the prior work in diversifying image sets of people focuses narrowly either on skin tone or perceived gender expression attributes of people. This baseline is used to understand the benefit of our more generalized person representation which does not list and classify people related attributes. For this baseline, a pretrained skin tone classifier and perceived gender expression classifier are used similar to those in \citet{muse_study}. We turn each signal into a simple continuous float, and produce a two dimensional ``skin tone + percevied gender expression'' representation. 
    \item \emph{CoCa Raw Embedding} To understand the added value of the text-guiding and human-alignment steps, another baseline is the original raw CoCa vision tower image embedding used as a starting point for PATHS \cite{yu2022coca}, which is expected to diversify broadly over all visual features.
    \item \emph{Text-derived Person-Diversity Representation} This baseline, only consisting of the text guiding step (Section \ref{subsec:text-guided}) on the original CoCa image embedding projected to the text-guided people-diversity representation, is used to show the value of the additional human-alignment step (Section \ref{subsec:human-rated-triplets}).
    \item \emph{Perception-aligned Person-Diversity Representation} Similarly, this baseline, created by only performing the perception-alignment step (Section \ref{subsec:human-rated-triplets}) on the original CoCa image embedding is used to demonstrate the value of the text-guiding step (Section \ref{subsec:text-guided}). Here the perception alignment learns a $1408 \times 12$ adapter matrix.
\end{itemize}

\subsection{Evaluation metric: Annotated End-to-End Diversity} \label{sec:eval_rate_e2e}

To evaluate the quality of the representation, 
4 annotators, one per region, from a pool of 8 human annotators across four regions (Brazil, India, Ghana, and Philippines) were asked to annotate whether the top 9 images created by a method were more diverse from a person-centric socio-cultural perspective than the undiversified top 9 images. The undiversified set of 9 images are collected by ranking according to the relevance computation alone. Not every pair of methods were annotated to minimize evaluation annotation. Details about annotations can be found in Appendix~\ref{app:human_annotations}.

The annotators either rate the diversification method as yielded an image set ``more diverse'', ``equivalently diverse'' or ``less diverse'' than the undiversified set.
The main metric for evaluating the diversification is the net change in diversity; the net percentage of queries that have more diverse results (the percentage of sets rated more diverse - percentage rated less, or, equivalently, an aggregate score of +1 for more diverse, -1 for less, and 0 for neutral ratings).

\section{Results}\label{sec:results}

\subsection{PATHS Improves Diversity over Both Narrow and Broad Methods}

Tests on the contrasting Occupations and DPD datasets show that \ourmethodabbv enables more ranking diversity improvement than both narrow baselines that only use skin tone and perceived gender expression signals, and broad baselines that are not people-specific (Table \ref{tab:sxs}\footnote{While Figure~\ref{tab:sxs} only shows net diversity change for a diversity weighting of $\alpha=0.5$ against no diversification, full results for other diversity weights and a head-to-head comparison between \ourmethodabbv and the skin tone and perceived gender expression baseline are included in Appendix \ref{app:full_sxs}. }).
On Occupations, the task that tests whether a representation promotes people-specific diversity over all general visual diversity (penalizing overly-broad approaches), \ourmethodabbv beats all baseline methods by a large margin: $+57.4\%$, compared to $+31.1\%$ for the SkinTone + Gender Expression baseline, and $-30.6\%$ for the CoCa Embedding.
We highlight that this improvement comes without the addition of costly attribute labels on people images, which prior literature tends to rely on to diversify more narrowly along skin tone and perceived gender.
On the Diverse People Dataset, created to test whether a representation can diversify across many types of people diversity (penalizing overly-narrow approaches), \ourmethodabbv also achieves among the best results: a $+53.6\% \pm 3.4\% $ diversity increase which is higher than the $+42.1\% \pm 3.0\% $  for the SkinTone + Gender Expression baseline, and close to the CoCa Embedding with $+55.6\% \pm 3.2\%$. Overall, \ourmethodabbv performs consistently well across both the "low available diversity" retrieval setting replicated by the Occupations dataset, and the "high available diversity" retrieval setting replicated by the Diverse People Dataset. 
An example of the end diversification achieved by \ourmethodabbv is depicted in Figures \ref{fig:qualitative_paths_only} and \ref{fig:qualitative}, where \ourmethodabbv promotes photos of traditional Chinese attire, and a woman in a wheelchair.

\newcolumntype{P}[1]{>{\centering\arraybackslash}p{#1}}
\begin{wraptable}{r}{0.55\textwidth}
\centering
\vspace{-10pt}
\begin{tabular}{m{0.2\textwidth} P{0.08\textwidth} P{0.08\textwidth} P{0.08\textwidth}}
\multicolumn{4}{l}{\textbf{Annotators prioritize gender expression over skin tone.} } \\
& Image A & Image B & Image C  \\ \toprule
Gender Expression  &  $G_1$ &   $G_1$ & $\mathbf{G_2}$ \\ 
Skin Tone  &  $\mathbf{S_1}$  &   $S_2$ &   $S_2$\\ 
\midrule
Annotators' Consensus on Different Image &  $11$ of $100$  &   $8$ of $100$& \underline{$\mathbf{81}$ of $100$}\\ 

\bottomrule
\end{tabular}
\caption{Given three different images, human annotators were asked to identify which image is most different from the other two. The three images are similar in most ways, but Image A has a different skin tone color from the other two images, and Image C has a different gender expression from the other two images. Both perceived gender expression and skin tone were determined by classifiers~\citep{muse_study}. This experiment was performed 100 times with 100 different such image triplets. We found that human annotators overwhelmingly choose Image C as the most different: that is, across a wide range of scenarios, they tend to believe that a different gender expression makes a person more different than a different skin tone. More analysis in Appendix \ref{appendix:analysis_of_gender_preference}.
}
\label{tab:rater_gender_over_skintone}
\vspace{-25pt}
\end{wraptable}

The CoCa embedding has starkly different diversity impacts on the two datasets: $-30.6\%$ on Occupations and $+55.6\%$ on Diverse People. This makes sense in context of each dataset. The Occupations dataset contain a large number of images which are visually diverse, but not demographically or culturally diverse. Here the lack of people-specificity in CoCa embeddings surfaces visually diverse, but not people-diverse images which reduces overall people diversity according to the human annotators.
Diverse People Dataset, however, only contains images where a person occupies most of the image pixels, and it features many attributes of people diversity. Broad visual diversity correlates very strongly with people diversity due to the nature of this dataset.

Lastly, each of the two stages of \ourmethodabbv alone --- the text-derived only representation, and the perception-aligned only representation --- do not perform as well as \ourmethodabbv. The perception-aligned person diversity representation has particularly poor performance, and possible reasons for this is discussed in Section~\ref{sec:discussion}.

\subsection{Human perception case studies}\label{subsec:complex_diversity}

Past research often assumes that changes in perceived gender expression and skin tone affect perceived diversity equally \citep{karako2018using,silva2023representation,celis2018fair}.
We can interrogate this assumption using collected human perception data, showing that counter to this, annotators may believe changes in gender expression are more are more salient to people diversity than changes in skin tone (Table \ref{tab:rater_gender_over_skintone}). 
\begin{wraptable}[33]{R}{0.55\textwidth}

    \centering
    \setlength{\tabcolsep}{1pt}
    \begin{tabular}{lccc}
          &  
        \begin{minipage}{15mm}
          \includegraphics[width=15mm]{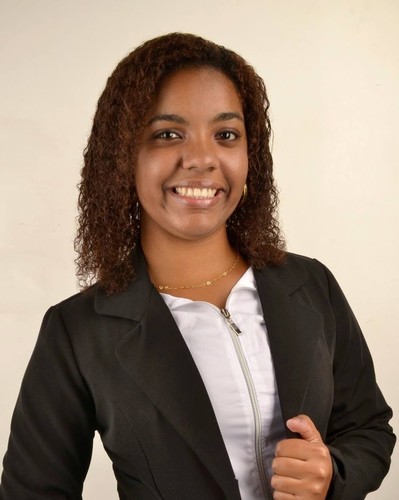}
        \end{minipage}
          &  
        \begin{minipage}{15mm}
          \includegraphics[width=15mm]{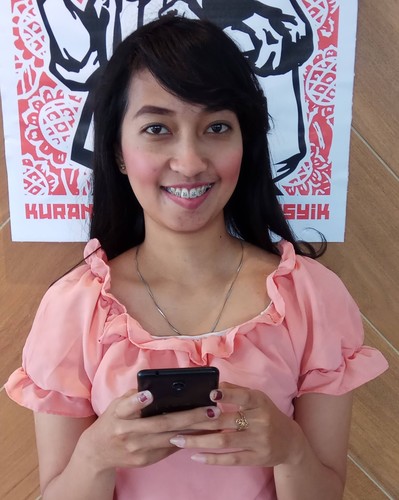}
        \end{minipage}
          &  
        \begin{minipage}{15mm}
          \includegraphics[width=15mm]{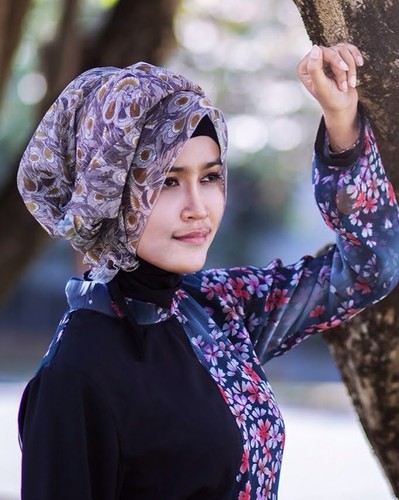}
        \end{minipage}
         \\
         \toprule
         Human Annotated & &  & \\
         Most Different & 1 of 4 & 0 of 4 & \underline{\textbf{3 of 4}}\\
         \midrule
         \multicolumn{4}{c}{Diversity Boost Given to Each Image During Ranking.}\\
         \multicolumn{4}{c}{Most boosted image is bolded.}\\
         \midrule
         Raw CoCa & $\textbf{2.863}$ & $1.532$ & $1.989$ \\
         SkinTone + Gender & $\textbf{0.053}$ & $-0.517$ & $-0.189$ \\
         Perception-Aligned Only & $\textbf{1.623}$ & $1.157$ & $0.620$ \\
         Text-Derived Only & $\textbf{1.714}$ & $-0.926$ & $-0.600$ \\
         PATHS & $0.508$ & $0.008$ & $\underline{\textbf{1.211}}$ \\
         \bottomrule
    \end{tabular}
    
    \vspace{0.2cm}
    
    \begin{tabular}{lccc}
          &  
        \begin{minipage}{15mm}
          \includegraphics[width=15mm]{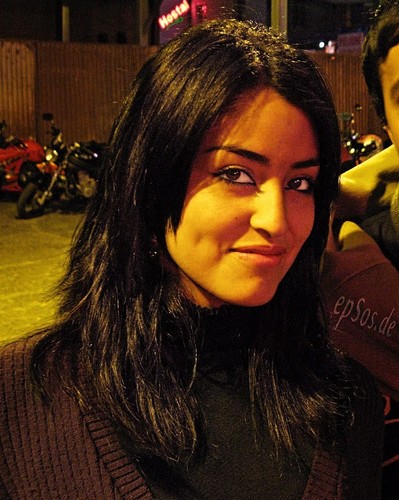}
        \end{minipage}
          &  
        \begin{minipage}{15mm}
          \includegraphics[width=15mm]{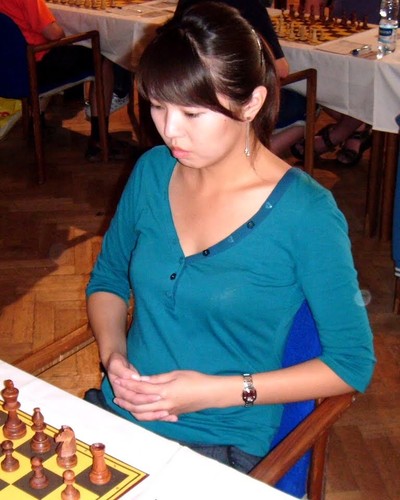}
        \end{minipage}
          &  
        \begin{minipage}{15mm}
          \includegraphics[width=15mm]{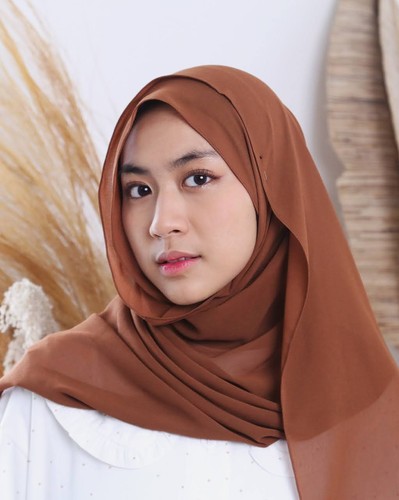}
        \end{minipage}
         \\
         \toprule
         Human Annotated & &  & \\
         Most Different & 0 of 4 & 0 of 4 & \underline{\textbf{4 of 4}}\\
         \midrule
         \multicolumn{4}{c}{Diversity Boost Given to Each Image During Ranking.}\\
         \multicolumn{4}{c}{Most boosted image is bolded.}\\
         \midrule
         Raw CoCa & $\textbf{0.214}$ & $-0.887$ & $-0.859$ \\
         SkinTone + Gender & $\textbf{-0.389}$ & $-0.659$ & $-0.730$ \\
         Perception-Aligned Only& $\textbf{-0.281}$ & $-0.576$ & $-0.422$ \\
         Text-Derived Only& $\textbf{0.518}$ & $-0.460$ & $0.380$ \\
         PATHS & $-1.231$ & $-1.306$ & $\underline{\textbf{-1.029}}$ \\
         \bottomrule
    \end{tabular} 
    
    \caption{Human annotators overwhelmingly believed wearing a hijab (right column) made a photo more different than having a different skin tone (left column). A diversity method that captures this human perception should boost the photo with the hijab the most. Only \ourmethodabbv does so. (Boost is Marginal Diversity in Alg. \ref{alg:mmr}.
    See App. \ref{app:image-attributions} for licences, attributions, and uncropped versions for these images.)}
    \label{tab:case_study_hijabs_1}
\end{wraptable}

Furthermore, this work finds that elements of cultural attire, such as the hijab, have an even greater impact on perceived diversity. In a study where 12 image triplets each contained one person wearing a hijab and two in Western attire, annotators across all four regions consistently rated the image with the individual wearing a hijab as the most different (methodology in Section \ref{subsubsec:get_human_triplet_ratings}).

Following up on this hijab effect, with a case study involving two new triplets (Table \ref{tab:case_study_hijabs_1}). The triplets contained images of two women with the same skin tone (one wearing a hijab, one not), and a third image of a woman with a different skin tone and no hijab. Even in this case, annotators strongly indicated that the presence or absence of a hijab contributed more to perceived difference than skin tone. Interestingly, only the \ourmethodabbv diversification method accurately reflected this human preference, highlighting the importance of the perception-alignment step in our approach (and the sensitivity of this method to the cultural context of the annotators). Appendix \ref{appendix:case_study} presents a similar case study with cultural bridal attire, leading to the same conclusions.

\section{Discussion, Limitations, and Future work}\label{sec:discussion}
\paragraph{Observation.} 
\ourmethodabbv involves two steps: 1) a text-guided PCA approach to find an initial person-diversity representation, 2) fine-tuning with human annotations to align the representation to human perception.  The results show that the perception alignment step alone (Section \ref{subsec:human-rated-triplets}) is insufficient: without the text-guided step, it does not lead to substantial ranking diversity gains (Table \ref{tab:sxs}) and does not help to capture human diversity preferences around cultural attire in qualitative explorations (Tables \ref{tab:case_study_hijabs_1} and \ref{tab:case_study_brides}).
This was surprising as both the perception alignment only method, and \ourmethodabbv achieve similar test errors ($70.0\%$ and $66.4\%$ respectively) in the human perception fine-tuning (Section \ref{subsubsec:metric_learning}). 

We hypothesize three reasons why perception alignment by itself may not work well: sparse data, difficulty in selecting useful ``hard'' triplets, and the gap between pairwise difference judgements and set-level diversity judgements.
When skipping the text-guiding step, $\sim17k$ parameters (a $1408 \times 12$ matrix) from only $30k$ data points are learned, since one is learning a projection directly on the pretrained CoCa embedding. 
When using the text guiding PCA step, however, one only has to learn a $12 \times 12$ matrix using human preference data, and can  further regularize to be close to the identity matrix since the text-guided projection is already capable. This is more constrained problem space, and therefore easier to learn (more discussion on this point in Appendix \ref{app:triplet_learning}).

On hard triplet selection,  
\ourmethodabbv mirrors older approaches used for person identification: this older work also uses metric learning only after pretraining a model with some surrogate model. %
\citet{hermans2017defense} shows that metric learning alone can outperform other methods for person identification by selecting the most useful batch of ``hard triplets'' at each step. Given the importance of selecting the right triplets, perhaps a tighter feedback loop with human annotators to seek out the most useful ``hard triplets'' in each training step is needed for the the perception alignment only method to succeed. 
Lastly, in the three-in-a-row task, human raters compare three images to assess the most different one. In the final task, they assess the diversity of the image set as a whole. The triplet loss may be misaligned to the set-level diversity task and modeling set level judgement directly may help. %

\paragraph{Limitations.} Though we ensured geographic diversity amongst our annotators, as we only used 8 annotators, we could not diversify across multiple dimensions e.g. self-identified gender, sexual orientation, body shape, education.
Similarly, while the members of the focus group (Section~\ref{subsec:text-guided}) were experts in diversity, fairness, and responsible AI; and included diverse individuals belonging to the LGBTQ+ and BIPoC communities, they were primarily US-based with high levels of education. Both factors may lead to some important dimensions of diversity being ignored due to unconscious biases.

Both the Occupations dataset~\citep{celis2020implicit} and the Diverse People Dataset are US-centric. The Occupations dataset is derived from occupations defined by the US Bureau of Labor and Statistics~\citep{kay2015unequal}, and the Diverse People Dataset was created using search queries from people in the US. Future work should expand these sets to be more globally inclusive.

A limited amount of training data that was sent for annotation (30k). A larger training set could lead to improvements in the expressiveness and alignment of the person-diversity representation space. That being said, even with a larger collection of annotations, this annotation method would need significantly fewer annotations those needed to produce individual attribute classifiers used in the current diversification methods.

We were unable to fully reduce our test error on the perception-alignment finetuning step (See Appendix \ref{app:triplet_learning}), meaning  we did not fully align to all perception preferences available, and better modeling methods may be needed.

\paragraph{Future Work.} As previous research has noted, annotations for subjective tasks are non-trivial to collect~\cite{schumann2023consensus,goyal2022toxicity,aroyo2023reasonable,diaz2022crowd}. One approach involves adopting model architectures targeting individual annotators instead of overall consensus estimate~\cite{chou2019every, davani2022dealing,  hayat2022modeling}. While this could lead to confirmation bias bubbles~\cite{ling2020confirmation}, it is an interesting direction of research to retrieve a diverse set of images satisfying the user's notion of diversity or the user's identity.
Another line of work could consider modeling diversity through exploring properties of image \textit{sets}, whereas our paper which mainly considers image-to-image similarity (e.g. triplets, MMR). Considering image sets as a first-class element may capture complex diversity preferences that our method cannot (Section~\ref{subsec:complex_diversity}). In addition, this paper focuses primarily on image retrieval and ranking systems. In the future this work could be adapted to generated images of people~\cite{lee2023aligning,friedrich2023fair,cho2023dall}.

\section{Conclusion}\label{sec:conclusion}

Diversely ranking images of people is a challenging task, one that requires striking a balance in capturing the many attributes of people-diversity, while avoiding attributes of non-people visual diversity. Our new method, \ourmethodabbv, accomplishes this balance better than past methods that narrowly focus on predefined people attributes, and better than methods that broadly diversify on all attributes of visual diversity. We find that our perception-alignment step, when combined with structure from the text-image space, also enables us to emphasize the types of people diversity that humans find to be most important (e.g. prioritizing salient cultural attire over differences in skin tone). There is still room for substantial future work. Nevertheless, this method will help enable ranking systems to embody richer, more flexible notions of visual people diversity by alleviating the need to prescriptively list and taxonomize people attributes or annotate photos of people.

\bibliographystyle{ACM-Reference-Format}
\bibliography{references}

\newpage
\clearpage
\appendix
\section{Qualitative Evals}
\begin{figure*}
    \centering
    \begin{subfigure}[b]{0.3\textwidth}
    \centering
    \includegraphics[width=\linewidth]{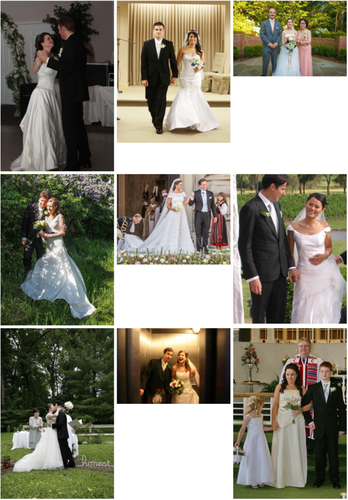}
    \caption{No Diversification}
    \label{fig:baseline_sxs}
    \end{subfigure}\hfill
    \begin{subfigure}[b]{0.3\textwidth}
    \centering
    \includegraphics[width=\linewidth]{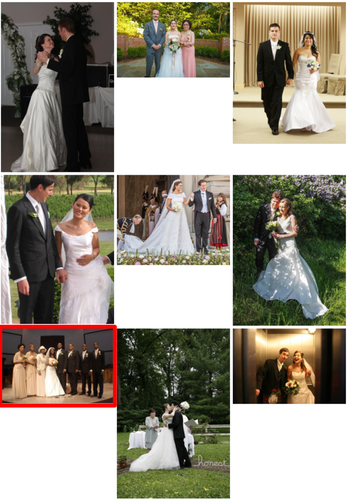}
    \caption{SkinTone + Gender Expression Baseline}
    \label{fig:baseline_sxs}
    \end{subfigure}\hfill
    \begin{subfigure}[b]{0.3\textwidth}
    \centering
    \includegraphics[width=\linewidth]{figures/side_by_sides/PATHS.png}
    \caption{PATHS}
    \label{fig:baseline_sxs}
    \end{subfigure}\hfill
    \caption{An example of image retrieval results for the query of ``bride'' over three different methods, No diversification baseline, SkinTone + Gender Expression Baseline, and PATHS.
    PATHS promotes two images outlined in red: a photo of a bride in traditionally Chinese cultural attire, and a photo of a disabled bride. SkinTone + Gender Expression Baseline promotes an image of a black bride surrounded by a wedding party. Here, we see that the gender component of this baseline creates an odd artifact: surface images that also have more men for the query ``bride.''}
    \label{fig:qualitative}
\end{figure*}

Figure \ref{fig:qualitative} depicts end to end ranking results for three treatments for the query ``bride'': no diversification, the skin tone and gender expression baseline, and PATHS. PATHS promotes two images: a photo of a bride in different cultural attire, and a photo of a disabled bride. The skin tone and gender expression baseline promotes an image of a black bride surrounded by a wedding party. Here, we see that the gender component of this baseline creates an odd artifact and causes us to surface images that also have more men for the query ``bride.''

\section{Results on other alpha values for diversity. Results for head to head comparison of skin tone + gender against PATHS.} \label{app:full_sxs}

\begin{figure*}
    \centering
    \begin{subfigure}[b]{.99\textwidth}
    \centering
    \includegraphics[width=\linewidth]{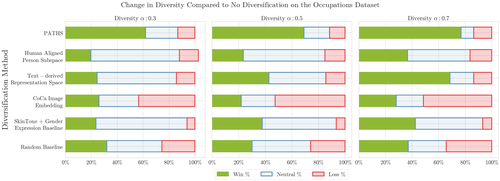}
    \label{fig:sxs_bar_chart}
    \end{subfigure}\hfill
    
    \begin{subfigure}[b]{0.99\textwidth}
    \centering
    \includegraphics[width=\linewidth]{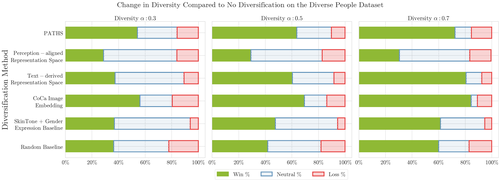}
    \label{fig:sxs_bar_chart}
    \end{subfigure}\hfill
    \caption{Full SxS results for all methods against the undiversified set, on both datasets for $\alpha=[0.3, 0.5, 0.7]$. "Wins" are where the diversification method's side was rated as more diverse, "neutral" was both sides were rated equally diverse, and "loss" is where the diversification method's side was rated as less diverse.}
    \label{fig:sxs_bar_chart_both}
\end{figure*}

Table \ref{tab:sxs_all_alpha} contains the increase in diversity for all methods, across both datasets, for all settings of $\alpha=[0.3, 0.5, 0.7]$. The text-derived representations, both text-derived only and PATHS, lose some of the people-related information from CoCa (Table \ref{tab:linear_probe_evals}). As such, they perform similarly or slightly worse than the raw CoCa in the Diverse People task, but still much better than the SkinTone + Gender or Random baselines that do not have information about other types of people diversity. In general, looking across results, \ourmethodabbv most consistently increases the diversity by a large amount (always >38\% increase). Figure \ref{fig:sxs_bar_chart_both} shows the full breakdown of ratings where "wins" are where the diversification method's side was rated as more diverse, "neutral" was both sides were rated equally diverse, and "loss" is where the diversification method's side was rated as less diverse. Table \ref{tab:direct_comparisons} has the head to head comparisons of \ourmethodabbv against the Skin tone + gender expression baseline and against our text-derived representation space. \ourmethodabbv overall increases diversity over the skin tone and gender expression baseline. The gap between \ourmethodabbv and the text-derived representation space is smaller, and more inconsistent, with \ourmethodabbv generally scoring slightly better.

\begin{table*}[]
    \centering
    \setlength{\tabcolsep}{3pt}
    \begin{tabular}{|l|r|r|r|}
        \toprule
         & \multicolumn{3}{c|}{Net Diversity Improvement ($\uparrow$ is better) for $\mathbf{\alpha=0.3}$}\\
        Diversification & \multicolumn{3}{c|}{(\% Queries Improved - \% Queries Worsened) }\\
        \cline{2-4}
        Treatment & Occupations Dataset & Diverse People Dataset & Avg. Across Datasets\\
        \hline
        Random Baseline & $6.6\%$ &  $14.3\%$ &   $14.5\%$\\
        SkinTone + Gender Expression Baseline & $17.5\%$ & $31.0\%$ & $24.3\%$ \\
        CoCa Image Embedding & $-17.5\%$ & $36.6\%$ & $9.6\%$\\
        \midrule
        \multicolumn{4}{l}{\textbf{\textit{Our Methods:}}} \\
        \midrule
        Text-derived Representation Space &  $10.2\%$ & $26.8\%$ & $18.5\%$\\
        Perception-aligned Representation Space & $5.2\%$ & $12.8\%$ & $9.0\%$ \\
        \ourmethodabbv & \underline{$\mathbf{48.7\%}$} & \underline{$\mathbf{38.4\%}$} & \underline{$\mathbf{43.6\%}$}\\
        \bottomrule
    \end{tabular}
    \setlength{\tabcolsep}{3pt}
    \begin{tabular}{|l|r|r|r|}
        \toprule
        \toprule
         & \multicolumn{3}{c|}{For $\mathbf{\alpha=0.5}$}\\
        Diversification & \multicolumn{3}{c|}{}\\
        \cline{2-4}
        Treatment & Occupations Dataset & Diverse People Dataset & Avg. Across Datasets\\
        \hline
        Random Baseline & $4.2\% \pm 3.8\%$ &  $23.8\% \pm 3.7\%$ &   $14.0\% \pm 5.3\%$\\
        SkinTone + Gender Expression Baseline & $31.2\% \pm 3.0\%$ & $42.1\% \pm 3.0\%$ & $36.6\% \pm 4.2\%$ \\
        CoCa Image Embedding & $-30.5\% \pm 4.1\%$ & \underline{$\mathbf{55.6\%\pm 3.6\%}$} & $12.5\% \pm 5.5\%$\\
        \midrule
        \multicolumn{4}{l}{\textbf{\textit{Our Methods:}}} \\
        \midrule
        Text-derived Representation Space &  $28.2\% \pm 3.6\%$ & \underline{$\mathbf{52.2\% \pm 3.2\%}$} & $40.2\% \pm 4.8\%$\\
        Perception-aligned Representation Space & $8.4\% \pm 3.2\%$ & $12.0\% \pm 3.4\%$ & $10.2\% \pm 4.6\%$ \\
        \ourmethodabbv & \underline{$\mathbf{57.9\% \pm 3.5\%}$} & \underline{$\mathbf{53.9\% \pm 3.4\%}$} & \underline{$\mathbf{55.9\% \pm 4.9\%}$}\\
        \bottomrule
    \end{tabular}
    \setlength{\tabcolsep}{3pt}
    \begin{tabular}{|l|r|r|r|}
        \toprule
        \toprule
         & \multicolumn{3}{c|}{For $\mathbf{\alpha=0.7}$}\\
        Diversification & \multicolumn{3}{c|}{}\\
        \cline{2-4}
        Treatment & Occupations Dataset & Diverse People Dataset & Avg. Across Datasets\\
        \hline
        Random Baseline & $2.9\%$ &  $43.0\%$ &   $23.0\% $\\
        SkinTone + Gender Expression Baseline & $35.7\%$ & $56.1\%$ & $45.9\%$ \\
        CoCa Image Embedding & $-23.3\%$ & \underline{$\mathbf{73.8\%}$} & $25.3\%$\\
        \midrule
        \multicolumn{4}{l}{\textbf{\textit{Our Methods:}}} \\
        \midrule
        Text-derived Representation Space &  $54.9\%$ & \underline{$\mathbf{73.3\%}$} & \underline{$\mathbf{64.1\%}$}\\
        Perception-aligned Representation Space & $20.3\%$ & $14.5\%$ & $17.4\%$ \\
        \ourmethodabbv & \underline{$\mathbf{63.3\%}$} & $57.1\%$ & $60.2\%$\\
        \bottomrule
    \end{tabular}
    \caption{Across both datasets, at $\alpha=[0.3, 0.5]$ our \ourmethodabbv has consistently the best impact on net diversity over the undiversified set. It is vastly better than other methods on the Occupations dataset, performs best on the Diverse People dataset for $\alpha=0.3$, and a very close second for $\alpha=0.5$. The CoCa Embedding performs starkly differently on both datasets for all values of $\alpha$. CoCa is a non-people specific general visual embedding: it does very poorly Occupations, the dataset that tests whether we promote people diversity specifically. It does very well on Diverse People, where most images are primarily focused on a person. There is a noticeable gap in performance between \ourmethodabbv and the CoCa Image embedding and the text-derived representation space for $\alpha=0.7$, so the method that achieves best performance when averaged across both datasets is the text-derived representation space. Overall, however, \ourmethodabbv most consistently performs well. 
    95\% confidence intervals are shown.}
    \label{tab:sxs_all_alpha}
    \vspace{-20pt}
\end{table*}

\begin{table*}[]
    \centering
    \setlength{\tabcolsep}{3pt}
    \begin{tabular}{|l|r|r|r|}
        \toprule
         & \multicolumn{3}{c|}{Net Diversity Improvement ($\uparrow$ is better) for $\mathbf{\alpha=0.3}$}\\
        Comparison & \multicolumn{3}{c|}{(\% Queries Improved - \% Queries Worsened) }\\
        \cline{2-4}
        Treatment & Occupations Dataset & Diverse People Dataset & Avg. Across Datasets\\
        \hline
        SkinTone + Gender Expression Baseline  & $38.7\%$ & $9.1\%$ & $23.9\%$ \\
        Text-derived Representation Space & $12.1\%$ & $-3.8\%$ & $4.2\%$\\
        \bottomrule
    \end{tabular}
    \begin{tabular}{|l|r|r|r|}
        \toprule
         & \multicolumn{3}{c|}{Net Diversity Improvement ($\uparrow$ is better) for $\mathbf{\alpha=0.5}$}\\
        Comparison & \multicolumn{3}{c|}{(\% Queries Improved - \% Queries Worsened) }\\
        \cline{2-4}
        Treatment & Occupations Dataset & Diverse People Dataset & Avg. Across Datasets\\
        \hline
        SkinTone + Gender Expression Baseline & $33.5\%$ & $13.6\%$ & $23.6\%$ \\
        Text-derived Representation Space & $0.8\%$ & $1.3\%$ & $1.1\%$\\
        \bottomrule
    \end{tabular}
    \begin{tabular}{|l|r|r|r|}
        \toprule
         & \multicolumn{3}{c|}{Net Diversity Improvement ($\uparrow$ is better) for $\mathbf{\alpha=0.7}$}\\
        Comparison & \multicolumn{3}{c|}{(\% Queries Improved - \% Queries Worsened) }\\
        \cline{2-4}
        Treatment & Occupations Dataset & Diverse People Dataset & Avg. Across Datasets\\
        \hline
        SkinTone + Gender Expression Baseline & $26.9\%$ & $18.9\%$ & $22.9\%$ \\
        Text-derived Representation Space & $2.7\%$ & $4.8\%$ & $3.8\%$\\
        \bottomrule
    \end{tabular}
    \caption{We directly compare \ourmethodabbv to The SkinTone + Gender Baseline as well as the Text-derived Representation Space. The Net Diversity improvements detail how much better PATHS is than the Comparison Treatment in the first column.}
    \label{tab:direct_comparisons}
    \vspace{-20pt}
\end{table*}

\section{Linear Probe Evaluations} \label{app:linear-probes}
\begin{figure}
    \centering
    \includegraphics[width=\linewidth]{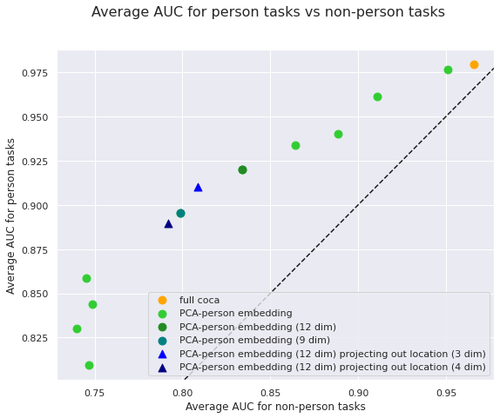}
    \caption{Results for the linear probe evaluations --- see Sec. \ref{subsec:text-guided}, Table \ref{tab:linear_probe_evals}, and Appendix \ref{app:linear-probes}.
    The X-axis shows the performance of a linear model trained on the embedding on non-person tasks, and the Y-axis shows this for person tasks (averaged AUC across tasks in both cases) --- we ideally want our model to perform well on person tasks, and poorly on non-person tasks.
    }
    \label{fig:linear_probes}
\end{figure}

Here we describe how we use a linear evaluation task as a heuristic for hyperparameter selection when choosing the values for $d_p$ (the number of PCA components we use when extracting the person diversity representation space initially from the CoCa embedding) and $d_b$ (the number of PCA components we use when projecting out information about the background from this initially extracted subspace) as defined in Sec. \ref{subsec:text-guided}. 
This also functions as a sanity check for the quality of the person-diversity representation.
We design a set of people-related and non-people related binary image classification tasks.
Each task is defined by a set of $q \geq 2$ queries (e.g. {``female doctor'', ``male doctor''} or {``American dancer'', ``Indian dancer'', ``Ghanaian dancer''}, {``walking indoors'', ``walking outdoors''}) --- for each of these queries we scrape a set of images from Google Image Search; the task is then a $q$-way classification problem, of distinguishing the $q$ sets of images from each other.
For each query set, we have some expectation around what the discriminative signal will be: for instance, in the ``female/male doctor'' task we expect the classification model will have to pick up on a signal around perceived gender expression in order to perform well, and in the ``American/Indian/Ghanaian dancer'' task, we expect the relevant signal will have something to do with cultural presentation.
In contrast, we expect the discriminative signal in the ``walking indoors/outdoors'' task will be unrelated to people --- we would call this a non-people-related task, and the tasks described previously to be people-related tasks
\footnote{Note that this does not mean we expect to be doing effective classification of percevied gender expression or culture per se; rather, we use this as a quick evaluation heuristic to estimate if the people-diversity representation responds more to signals which are likely people-related than those which are likely non-people-related.}.

For each task, we learn a simple linear layer on top of the existing people-diversity representations. Ideally, the people-diversity representation performs well on the person-related tasks and poorly on the non-person tasks.
Table~\ref{tab:linear_probe_evals} summarizes the results of these evaluations for the candidate people-diversity representation.
We ran a sweep across values of $d_p, d_b$ and compared performance on these linear evaluations, and chose $d_p = 12, d_b = 3$, i.e.  a 12 dimensional PCA-derived people-diversity representation with 3 dimensions of the ``person-background'' subspace projected out, which seemed to present the best tradeoff between high performance on the people tasks, low performance on non-people tasks, and small representation size.
The final candidate is a 12 dimensional PCA-derived people-diversity representation with 3 dimensions of the ``person-background'' subspace projected out.
The result is $P$, a linear transformation of the original CoCa image embedding that projects to a ``text-derived person diversity subspace.''

In Fig. \ref{fig:linear_probes}, we show the results of our hyperparameter sweep over $d_p$ and $d_b$.
Our evaluation method is to train a linear model to predict which image corresponds to which query (e.g. ``female doctor'' vs ``male doctor''). %
The X-axis shows the performance of a linear model trained on the embedding on non-person tasks, and the Y-axis shows this for person tasks (averaged across tasks in both cases) --- we ideally want our model to perform well on person tasks, and poorly on non-person tasks.
For example, in the top right hand corner is the point representing the full CoCa embedding, which has high performance on both person and non-person tasks --- all other options are worse on both task groups.
We see that there are many viable choices which sit on various points across the tradeoff between these two, represented by various circles on the plot.
We show with triangles and darker circles the options we decided presented the best points along this tradeoff --- we picked these since they seemed to sit at an ``elbow'' point where the tradeoff was slightly better, as shown by distance from the black diagonal (representing equal performance on the two task groups).

\section{Learning from Human Annotations on Triplets} \label{app:triplet_learning}
In this section we'll present and discuss the training set-up to incorporate human perception results, in the ``perception-alignment'' phase described in Section \ref{subsubsec:metric_learning}. 

We learned a simple linear matrix transformation in all cases, using all $30k$ human annotations shuffled together. The batch size was $1,000$, number of steps $60,000$, with a validation step performed every $600$ steps. We used an Adam Optimizer with learning rate $0.0001$ and momentum $0.9$. The margin on the triplet loss ($\beta$ in Equation \ref{eqn:triplet_loss}) was set to $0$. 

We compute error on a per triplet basis: a triplet is correct, iff all the desired relative similarities (There are 2-3 per triplet as described in Table \ref{tab:diff_to_sim}) are correctly enforced. The starting and ending test errors for the perception alignment only and \ourmethodabbv approaches are depicted in Table \ref{tab:triplet_learning_err}.

All approaches have the same (high) test error post-fine tuning. This is true even through the perception alignment only method starting with a much lower test error (This method starts by passing in the top 12 rows of the  original CoCa embedding). Surprisingly then, \ourmethodabbv clearly outperforms perception alignment alone on end-to-end diversity impact, as demonstrated in Figures \ref{fig:sxs_occ} and \ref{fig:sxs_pdcd}. It's clear that the test error on this individual triplet task correlates somewhat poorly with the end Diversity Impact, but it is unclear why. Likely, this task's error correlates poorly with end Diversity Impact simply perhaps because even this task's error is still so high at the end of fine-tuning. Furthermore to explain why the additive \ourmethodabbv performs so much worse than the projection approach (columns 3 and 2 of Table \ref{tab:triplet_learning_err} respectively), we hypothesize that with the relatively sparse data available, we're not able to learn many parameters.

\newcolumntype{R}[1]{>{\raggedleft\arraybackslash}p{#1}}
\begin{table*}[]
    \centering
    \begin{tabular}{R{0.18\textwidth}| l | l| l}
                  &  Perception Alignment Only  &  
                  \ourmethodabbv : Multiplication Version &
                  \ourmethodabbv : Additive Version\\
                  &  (No text-guiding step)  &  
                   \textbf{(Version used in Paper)} &
                   (Not used in Paper)\\
       \toprule
       Dimension of Learned Matrix, $M$ &  $1408 \times 12 $&  $12 \times 12 $ &  $1408 \times 12 $ \\
       How to compute embedding, $e(I,M)$ & 
       $\text{CoCa\_Embedding}(I) \cdot M $ & 
       $\text{CoCa\_Embedding}(I) \cdot P \cdot M$  & 
       $\text{CoCa\_Embedding}(I) \cdot M$ \\
       Matrix regularized against for L1, L2 Loss & 
        $\begin{pmatrix}\begin{matrix}
          I_{12 \times 12} \\
          0_{1396 \times 12}
        \end{matrix}\end{pmatrix}$
       & $I_{12 \times 12}$ & $P$ \\
       L1 Loss Weight & $(1408 \cdot 12)^{-1}$ &$(12 \cdot 12)^{-1}$ &$(1408 \cdot 12)^{-1}$\\
       L2 Loss Weight & $(1408 \cdot 12)^{-1}$ &$(12 \cdot 12)^{-1}$ &$(1408 \cdot 12)^{-1}$\\
      Test Error Before Fine-tuning & $74.6\%$ & $96.9\%$ & $96.9\%$\\
       Test Error After Fine-tuning &   $70.0\%$& $66.4\%$ & $78.1\%$\\
       \bottomrule
    \end{tabular}
    \caption{We summarize the approaches tried for human perception alignment. $M$ is the matrix being learned in this step. The first only does perception alignment only without any text-guiding step. The second and third fine-tune the output of the text guided step, $P$ (Section \ref{subsec:text-guided}). Recall the $P$ is the linear transformation of the original CoCa image embedding that projects to a people-diversity representation. The second column, is the version of \ourmethodabbv we use in the paper which learns a $12 \times 12$ projection we multiply $P$ with. The third method (unused due to the high test error rate) seeks to modify the individual params of $P$ itself. This additive approach does not do as well as the multiplicative approach of column 2. }
    \label{tab:triplet_learning_err}
    \vspace{-20pt}
\end{table*}

\section{Human Annotations}\label{app:human_annotations}
Research has shown the importance of annotator diversity when performing subjective annotation tasks~\cite{schumann2023consensus,goyal2022toxicity,diaz2022crowd}. Therefore, following \citet{schumann2023consensus}, we hired 8 annotators evenly spread across 4 regions: Brazil, Ghana, India, and the Philippines.

We first tested three methods of getting annotations to identify which method was best at measuring the general people diversity in images. The first method was pairwise annotation, in which annotators would be asked to view a target image and annotate the extent to which the paired image was different from the target image on a 3-point scale (see Appendix~\ref{app:human_annotations}). The second method we considered was triplet annotation, in which annotators would view a target image and two more images, Image A and Image B, and select which of these two images were more different from the target image (see Appendix~\ref{app:human_annotations}). The third method we considered was three-in-a-row annotation, in which annotators would view three images, Image A, Image B, and Image C, and select which of the three images most the most different of the set (see Appendix~\ref{app:human_annotations}). The image sets used to test the three methods were selected to include both easy examples, which had clear "correct" answers, and hard examples, where there is no clear "correct" answer. These two type of sets allow us to a) understand if the annotators are able to perform the task correctly and b) look at consensus across hard examples.

Of the three methods, we selected three-in-a-row for our model development. The three methods had similar amounts of ``accuracy'' --- agreement between annotators and our internal team --- but the three-in-a-row had the highest rater consensus and the highest rate of 100\% agreement between annotators. For the full details on our annotation method selection see the appendix~\ref{app:human_annotations}.

In our initial examination we sent a test set of 30 comparisons in each condition to the group of eight annotators, two from India, Brazil, Ghana, and Philippines. One limitation of our test was that annotators saw the same images, presented randomly, across the annotation conditions. In consideration of order effects, where practice might improve performance over time, we split the eight annotators into two groups of four with one annotator from each region. Group One saw three-in-a-row, then pairwise, then triplets, and Group Two saw triplet, pairwise, three-in-a-row. In determining which method to use we considered the relative accuracy of the methods and the consistency in annotation among annotators.

There were eighteen trials in which there were clear “right” answers as defined by the team of authors. Across methods we found a similar amount of accuracy for these items. In triplets and three-in-a-row annotators would select the correct image as most different. In pairwise the average difference in the paired image would be greater for the “right” answer. Across annotation methods there were similar rates of accuracy. Given these similar rates of accuracy (Table~\ref{tab:sensitivity_correctness}) we focused on consistency among annotators when selecting which method to use for model development.

\begin{table}[b]
\begin{tabular}{llll}
                             & Three-in-a-Row & Pairwise & Triplet \\ \hline
Group 1 & 15/18          & 16/18    & 18/18   \\
Group 2 & 18/18          & 17/18    & 17/18  
\end{tabular}
\caption{Number of correct answers per sensitivity task.}
\label{tab:sensitivity_correctness}
\end{table}

For consensus among annotators we considered the whole set of 30 test questions, 18 questions with “right” answers and 12 questions with no-clear answer when considering which method to use for annotation. Pairwise performed the worse when it came to consensus - in part because the method might be too sensitive to differences in how the annotators may use the 3-point scale. For example, in one pairwise case where expected there to be no differences, two outwardly similar feminine presenting women,  only two annotators across Group One and Group Two selected “No Different” when comparing the two images. Suggestive of an over-sensitive measure, five of the annotators selected “Somewhat Different” noting features like hair color or uniform color, and one annotator selected “Very Different” (although this may have been in error as they noted the two people had similar age and gender expressions). This also highlighted an issue of interpretability across different comparisons, in the example above an annotator rated both the top comparisons, two outwardly similar feminine presenting women, and the bottom comparison, one feminine presenting women and one masculine presenting man, as “somewhat different” similarly weighting the difference between uniform color and gender expression. Although this might have been one case of misuse of the annotation scale, we decided not to move forward with pairwise because of a greater number of potential issues with consensus, over-sensitivity, and the interoperability of the annotations.

When comparing triplet and three-in-a-row annotations we paid special attention to the order in which annotators used the various methods. Group One saw triplets last and Group Two saw three-in-a-row last, suggesting they had the most practice prior to completing these condition. Generally, we found similar rates of consensus across methods even when considering the order in which annotators were exposed to the various conditions (Table~\ref{tab:consensus}). However, Group Two’s three-in-a-row performance stood out. Not only did they have the highest rate of consensus on the test set, but the highest rate of 100\% agreement among the annotators. This was particularly interesting because three-in-a-row is a three-choice paradigm where triplets are a two-choice paradigm, suggesting that despite the higher chance for disagreement annotators are performing better on this more difficult task. Thus, we decided to leverage the three-in-a-row method of annotators for the general diversity ratings.

\begin{table}[]
\begin{tabular}{lll}
      & Group 1                              & Group 2                              \\
\hline
Triplet        & \textbf{*83.33\% {\small (saw 2nd)}} & 85\% {\small  (saw 1st) }             \\
Three-in-a-Row & 84.17\% {\small  (saw 1st) }          & \textbf{*90.83\% {\small (saw 2nd)}}
\end{tabular}
\caption{Percentage of consensus across sensitivity tasks where consensus is defined as the number of tasks with 100\% agreement between annotators for the image choice.}
\label{tab:consensus}
\vspace{-20pt}
\end{table}

\subsection{Human Annotation Templates}
\begin{figure*}
    \centering
    \includegraphics[width=\textwidth]{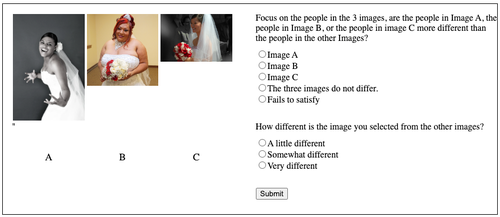}
    \caption{Three-in-a-row template}
    \label{fig:three_in_a_row}
\end{figure*}

\begin{figure*}
    \centering
    \includegraphics[width=\textwidth]{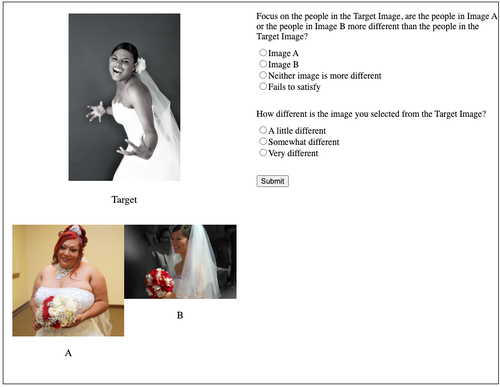}
    \caption{Triplet template}
    \label{fig:triplet}
\end{figure*}

\begin{figure*}
    \centering
    \includegraphics[width=\textwidth]{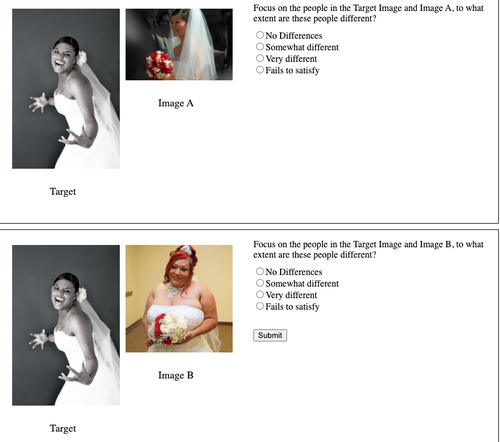}
    \caption{Pairwise template}
    \label{fig:pairwise}
\end{figure*}

\begin{figure*}
    \centering
    \includegraphics[width=\textwidth]{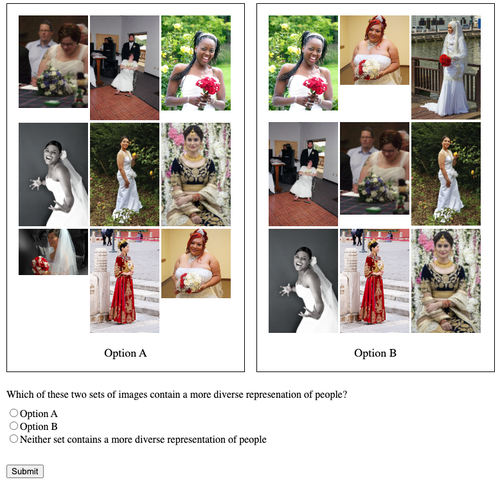}
    \caption{Side by side diversity template}
    \label{fig:side_by_side_template}
\end{figure*}

Human annotators used the following templates for annotation tasks.
\begin{itemize}
    \item Three-in-a-Row in Figure~\ref{fig:three_in_a_row}
    \item Triplet in Figure~\ref{fig:triplet}
    \item Pairwise in Figure~\ref{fig:pairwise}
    \item Side by side diversity evaluation in Figure~\ref{fig:side_by_side_template}
\end{itemize}

For the side-by-side diversity template, which set of images appeared on the left vs the right was randomized. 

\section{Additional Case Study: Cultural Wedding Dresses} \label{appendix:case_study}
This additional case study on photos of brides from different cultures highlights the type of diversity gains we see due to the human-alignment step of \ourmethodabbv. In Table \ref{tab:case_study_brides} we see that human annotators believe that a person wearing bridal attire from a different culture makes the person more different than a person of a different skin tone wearing bridal attire from the same culture. This human preference is captured in \ourmethodabbv, which has both text-guiding and human-aligning steps. Raw CoCa embedding also gets the second example right, though in the second example the backdrop and overall visual difference of the third image of the Chinese bride is also greatest --- i.e. it's unclear whether the raw CoCa gets the example right due to the people diversity or the backdrop and lighting diversity.

\begin{table}[h]

    \centering
    
    \begin{tabular}{lccc}
          &  
        \begin{minipage}{15mm}
          \includegraphics[width=15mm]{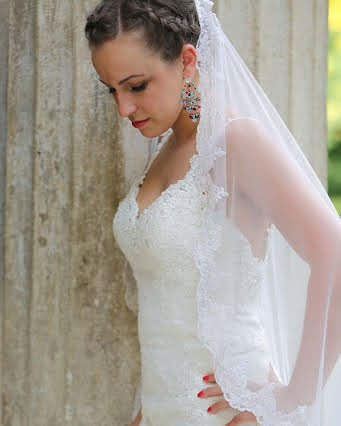}
        \end{minipage}
          &  
        \begin{minipage}{15mm}
          \includegraphics[width=15mm]{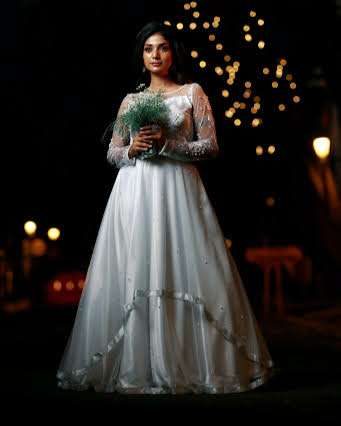}
        \end{minipage}
          &  
        \begin{minipage}{15mm}
          \includegraphics[width=15mm]{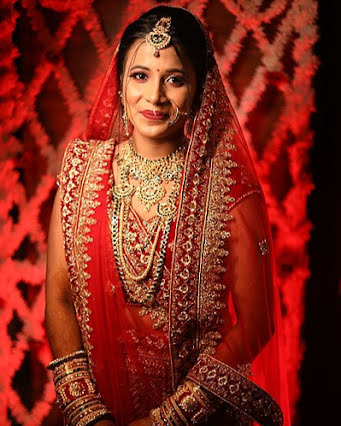}
        \end{minipage}
         \\
         \toprule
         Human Annotated & &  & \\
         Most Different & 0 of 4 & 0 of 4 & \underline{\textbf{4 of 4}}\\
         \midrule
         \multicolumn{4}{c}{Diversity Boost Given to Each Image During Ranking.}\\
         \multicolumn{4}{c}{Most boosted image is bolded.}\\
         \midrule
         Raw CoCa & $\textbf{-0.803}$ & $-0.836$ & $-0.813$ \\
         Human-Aligned & $\textbf{-0.295}$ & $-0.415$ & $-0.587$ \\
         SkinTone + Gender & $\textbf{-0.027}$ & $-0.584$ & $-0.592$ \\
         Text-Derived & $\textbf{1.638}$ & $-1.006$ & $-0.646$ \\
         PATHS & $-1.023$ & $-1.238$ & $\underline{\textbf{-0.878}}$ \\
         \bottomrule
    \end{tabular}
    
    \vspace{0.5cm}
    
    \begin{tabular}{lccc}
          &  
        \begin{minipage}{15mm}
          \includegraphics[width=15mm]{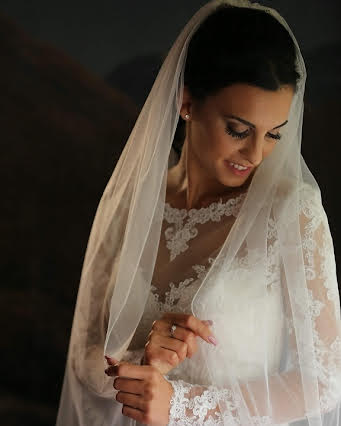}
        \end{minipage}
          &  
        \begin{minipage}{15mm}
          \includegraphics[width=15mm]{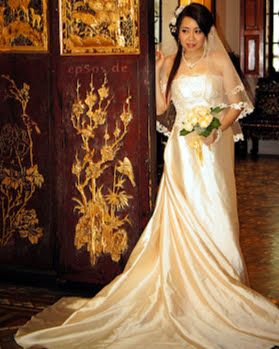}
        \end{minipage}
          &  
        \begin{minipage}{15mm}
          \includegraphics[width=15mm]{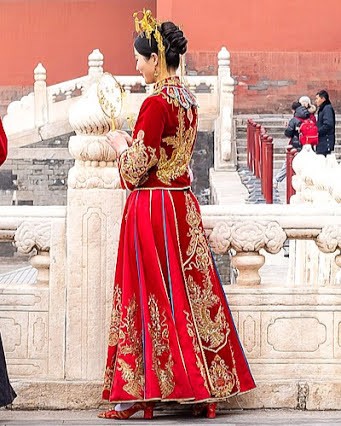}
        \end{minipage}
         \\
         \toprule
         Human Annotated & &  & \\
         Most Different & 0 of 4 & 0 of 4 & \underline{\textbf{4 of 4}}\\
         \midrule
         \multicolumn{4}{c}{Diversity Boost Given to Each Image During Ranking.}\\
         \multicolumn{4}{c}{Most boosted image is bolded.}\\
         \midrule
         Raw CoCa & $-0.178$ & $-0.001$ & $\underline{\textbf{0.174}}$ \\
         Human-Aligned & $-0.338$ & $\textbf{-0.269}$ &  $-0.454$ \\
         SkinTone + Gender & $\textbf{-1.005}$ & $-1.227$ & $-1.240$ \\
         Text-Derived & $\textbf{0.078}$ & $-1.773$ & $-1.222$ \\
         PATHS & $0.823$ & $0.470$ & $\underline{\textbf{2.166}}$ \\
         \bottomrule
    \end{tabular} 
    
    \caption{This figure presents two sets of triplet images we asked human annotators to annotate (None of these images were used in any other training or evaluation step.). The right two columns of images both have the same skin tone. The left two columns both wearing bridal attire from the same culture. In both cases, human annotators find the woman wearing bridal attire from a different culture to be the most different of the three photos. This preference is reflected in the \ourmethodabbv -- which is both text-derived and human-aligned. In the second example, the raw CoCa also gets the correct image. (Boost refers to the equation in row 6 of Algorithm \ref{alg:mmr}.) Attributions, licenses and uncropped images can be found in Appendix \ref{app:image-attributions}.}
    \label{tab:case_study_brides}
\end{table}

\section{Analysis of Human Annotator preference for Gender over Skin Tone.}\label{appendix:analysis_of_gender_preference}
For all of the 100 triplets, age was similar across the triplets to minimize the effect of other attributes. While we can't account for all other perceptual attributes being equivalent across the triplets, the trend towards the image with a different gender expression is strong, with 81 out of 100 triplets having the most different image selected as the one with different gender expression. Please see Figure~\ref{tab:case_study_brides} for a similar case study to that in Section~\ref{sec:discussion}.

\section{Image Attributions} \label{app:image-attributions}

Information for image attributions throughout the paper can be found in this section.

Images in Figure~\ref{fig:qualitative_paths_only}:
\begin{itemize}
    \item \url{https://c1.staticflickr.com/3/2506/4000554611_df3dcd69df_o.jpg} (Derek Hatfield, CC-BY 2.0)
    \item \url{https://en.wikipedia.org/wiki/File:A_bride_in_Xiuhe_dress_between_Taihedian_and_Zhonghedian_\%2820220218120518\%29.jpg} (CC-BY 4.0)
    \item \url{https://farm8.staticflickr.com/7099/7308570096_f97ede5dc3_o.jpg} (Jenny Kristoffer, CC-BY 2.0)
    \item \url{https://farm4.staticflickr.com/7139/7594113252_3ba7086f47_o.jpg} (lina smith, CC-BY 2.0)
    \item \url{https://c2.staticflickr.com/1/1/1191578_4f95f80d43_o.jpg} (April, CC-BY 2.0)
    \item \url{https://openverse.org/image/32839872-b4d6-40ba-9b0a-debe345df175?q=bride\%20wheelchair}, (sylvar, CC-BY 2.0)
    \item \url{https://c1.staticflickr.com/6/5581/14930736036_63ec66ab29_o.jpg} (mistywordpower, CC-BY 2.0)
    \item \url{https://c6.staticflickr.com/8/7352/8994126007_d407c99ee7_o.jpg} (Bengt Nyman, CC-BY 2.0)
    \item \url{https://c2.staticflickr.com/1/255/18531054670_565d3e0d99_o.jpg} (Scott Riggle, CC-BY 2.0)
\end{itemize}

\begin{figure}
    \centering
    \includegraphics[height=0.7\linewidth]{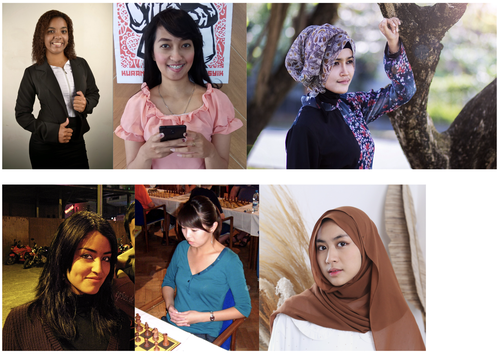}
    \caption{Uncropped images (as they were presented to human annotators) for the images used in Table \ref{tab:case_study_hijabs_1}. Attribution and licenses for these images are in Appendix \ref{app:image-attributions}.}
    \label{fig:hijab_case_study_uncropped}
\end{figure}

Images in Table ~\ref{tab:case_study_hijabs_1} (L to R):

\begin{itemize}
    \item Top row:
    \begin{itemize}
        \item \url{https://pxhere.com/en/photo/706058} (CC0)
        \item \url{https://openverse.org/image/88bf1e38-aeb1-4bf0-b929-11d21b9eb099} (Alex Neman, CC BY-SA 4.0)
        \item \url{https://openverse.org/image/3e5a26b2-2508-4a23-8cec-c77813056762} (Ferian-pz, CC BY 2.0)
        \item Uncropped images as they were shown to human annotators can be found in in Figure \ref{fig:hijab_case_study_uncropped}.
    \end{itemize}
    \item Second row:
    \begin{itemize}
        \item \url{https://en.m.wikipedia.org/wiki/File:Beautiful_Latina_Woman_Smiling.jpg} (epSos.de, CC BY 2.0)
        \item \url{https://commons.wikimedia.org/wiki/File:Dulamsuren_Yanjindulam_2010.JPG} (Wind87, public domain)
        \item \url{https://openverse.org/image/dd11cfe4-b35c-4264-8b7d-ea11632fdb26} (Dede Rifan, CC BY-SA 4.0)
        \item Uncropped images as they were shown to human annotators can be found in in Figure \ref{fig:hijab_case_study_uncropped}.
    \end{itemize}
\end{itemize}

Images in Figures~\ref{fig:qualitative}:
\begin{itemize}
    \item \url{https://c1.staticflickr.com/3/2506/4000554611_df3dcd69df_o.jpg} (Derek Hatfield, CC-BY 2.0)
    \item \url{https://farm4.staticflickr.com/7139/7594113252_3ba7086f47_o.jpg} (lina smith, CC-BY 2.0)
    \item \url{https://c2.staticflickr.com/1/255/18531054670_565d3e0d99_o.jpg} (Scott Riggle, CC-BY 2.0)
    \item \url{https://farm8.staticflickr.com/7099/7308570096_f97ede5dc3_o.jpg} (Jenny Kristoffer, CC-BY 2.0)
    \item \url{https://c6.staticflickr.com/8/7352/8994126007_d407c99ee7_o.jpg} (Bengt Nyman, CC-BY 2.0)
    \item \url{https://c2.staticflickr.com/1/1/1191578_4f95f80d43_o.jpg} (April, CC-BY 2.0)
    \item \url{https://c1.staticflickr.com/6/5581/14930736036_63ec66ab29_o.jpg} (mistywordpower, CC-BY 2.0)
    \item \url{https://c8.staticflickr.com/6/5227/5610128766_5d230cedc4_o.jpg} (Lynn B, CC-BY 2.0)
    \item \url{https://c1.staticflickr.com/5/4072/4677014642_e1a258b40a_o.jpg} (Cliff, CC-BY 2.0)
    \item \url{https://c1.staticflickr.com/3/2506/4000554611_df3dcd69df_o.jpg} (Derek Hatfield, CC-BY 2.0)
    \item \url{https://c2.staticflickr.com/1/255/18531054670_565d3e0d99_o.jpg} (Scott Riggle, CC-BY 2.0)
    \item \url{https://farm4.staticflickr.com/7139/7594113252_3ba7086f47_o.jpg} (lina smith, CC-BY 2.0)
    \item \url{https://c2.staticflickr.com/1/1/1191578_4f95f80d43_o.jpg} (April, CC-BY 2.0)
    \item \url{https://c6.staticflickr.com/8/7352/8994126007_d407c99ee7_o.jpg} (Bengt Nyman, CC-BY 2.0)
    \item \url{https://farm8.staticflickr.com/7099/7308570096_f97ede5dc3_o.jpg} (Jenny Kristoffer, CC-BY 2.0)
    \item \url{https://c8.staticflickr.com/6/5255/5452907146_b2a55a2f35_o.jpg} (Jerald Jackson, CC-BY 2.0)
    \item \url{https://c1.staticflickr.com/6/5581/14930736036_63ec66ab29_o.jpg} (mistywordpower, CC-BY 2.0)
    \item \url{https://c8.staticflickr.com/6/5227/5610128766_5d230cedc4_o.jpg} (Lynn B, CC-BY 2.0)
    \item \url{https://c1.staticflickr.com/3/2506/4000554611_df3dcd69df_o.jpg} (Derek Hatfield, CC-BY 2.0)
    \item \url{https://en.wikipedia.org/wiki/File:A_bride_in_Xiuhe_dress_between_Taihedian_and_Zhonghedian_\%2820220218120518\%29.jpg} (CC-BY 4.0)
    \item \url{https://farm8.staticflickr.com/7099/7308570096_f97ede5dc3_o.jpg} (Jenny Kristoffer, CC-BY 2.0)
    \item \url{https://farm4.staticflickr.com/7139/7594113252_3ba7086f47_o.jpg} (lina smith, CC-BY 2.0)
    \item \url{https://c2.staticflickr.com/1/1/1191578_4f95f80d43_o.jpg} (April, CC-BY 2.0)
    \item \url{https://openverse.org/image/32839872-b4d6-40ba-9b0a-debe345df175?q=bride\%20wheelchair}, (sylvar, CC-BY 2.0)
    \item \url{https://c1.staticflickr.com/6/5581/14930736036_63ec66ab29_o.jpg} (mistywordpower, CC-BY 2.0)
    \item \url{https://c6.staticflickr.com/8/7352/8994126007_d407c99ee7_o.jpg} (Bengt Nyman, CC-BY 2.0)
    \item \url{https://c2.staticflickr.com/1/255/18531054670_565d3e0d99_o.jpg} (Scott Riggle, CC-BY 2.0)
\end{itemize}

Images in Figures~\ref{fig:three_in_a_row}, \ref{fig:triplet}, and \ref{fig:pairwise}:
\begin{itemize}
    \item \url{https://openverse.org/image/08bc90f9-3b80-4835-b50a-5cc841e6be65?q=bride}. (CherieJPhotography, CC BY 2.0)
    \item \url{https://www.flickr.com/photos/taedc/14957841864} (Ted Eytan, CC BY-SA 2.0 DEED)
    \item \url{https://en.m.wikipedia.org/wiki/File:Bride_with_bouquet.jpg} (Sherry Main from USA, CC BY-SA 2.0 DEED)
\end{itemize}

Images in Figure~\ref{fig:side_by_side_template}:
\begin{itemize}
    \item \url{https://www.flickr.com/photos/dvanzuijlekom/27045165656/} (Dennis van Zuijlekom, CC BY-SA 2.0 DEED)
    \item \url{https://openverse.org/image/32839872-b4d6-40ba-9b0a-debe345df175?q=bride\%20wheelchair} (sylvar, CC BY 2.0)
    \item \url{https://www.publicdomainpictures.net/en/view-image.php?image=110301&picture=happy-wedding-day} (Geoff Doggett, CC0 Public Domain)
    \item \url{https://openverse.org/image/08bc90f9-3b80-4835-b50a-5cc841e6be65?q=bride}. (CherieJPhotography, CC BY 2.0)
    \item \url{https://openverse.org/image/ef05c64d-4272-4e56-9c5b-3f60f6d98174?q=tehran\%20bride} (Taymaz Valley, CC BY 2.0)
    \item \url{https://www.pxfuel.com/en/free-photo-xsbcj} (Free for commercial use, DMCA)
    \item \url{https://www.flickr.com/photos/taedc/14957841864} (Ted Eytan, CC BY-SA 2.0 DEED)
    \item \url{https://en.wikipedia.org/wiki/File:A_bride_in_Xiuhe_dress_between_Taihedian_and_Zhonghedian_\%2820220218120518\%29.jpg} (CC BY 4.0)
    \item \url{https://en.m.wikipedia.org/wiki/File:Bride_with_bouquet.jpg} (Sherry Main from USA, CC BY-SA 2.0 DEED)
\end{itemize}

\begin{figure}
    \centering
    \includegraphics[height=0.75\linewidth]{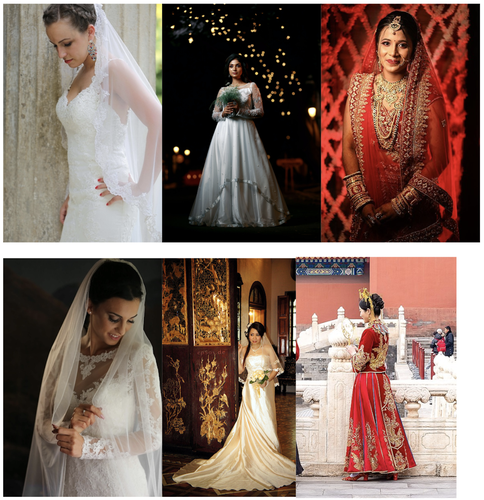}
    \caption{Uncropped images (as they were presented to human annotators) for the images used in Table \ref{tab:case_study_brides}. Attribution and licenses for these images are in Appendix \ref{app:image-attributions}.}
    \label{fig:bride_case_study_uncropped}
\end{figure}

Images in Table \ref{tab:case_study_brides}:
\begin{itemize}
    \item Top row (L to R):
    \begin{itemize}
        \item \url{https://pixnio.com/media/pretty-bride-wedding-dress-earrings-side-view} (Milivojevic, Free to use, CC0)
        \item \url{https://pxhere.com/en/photo/1671362} (Purest / 554 Images, public domain, CC0)
        \item \url{https://commons.wikimedia.org/wiki/File:Indian_Bride_dress_on_wedding_day.jpg} (AmanAgrahari01, Creative Commons Attribution-Share Alike 4.0 International)
        \item Uncropped images as they were shown to human annotators can be found in in Figure \ref{fig:bride_case_study_uncropped}.
    \end{itemize}
    \item Bottom row (L to R):
    \begin{itemize}
        \item \url{https://pixnio.com/media/photo-studio-wedding-dress-wedding-bride-posing} (Milivojevic, CC0)
        \item \url{https://commons.wikimedia.org/wiki/File:Beautiful_Chinese_Bride_in_White_Wedding_Dress.jpg} (epSos.de, Creative Commons Attribution 2.0 Generic)
        \item \url{https://en.m.wikipedia.org/wiki/File:A_bride_in_Xiuhe_dress_near_Xiehemen_\%2820220218110303\%29.jpg} (N509FZ, Creative Commons Attribution-Share Alike 4.0 International)
        \item Uncropped images as they were shown to human annotators can be found in in Figure \ref{fig:bride_case_study_uncropped}.
    \end{itemize}
\end{itemize}

\section{Diverse People Dataset (DPD) Data Card}\label{app:datacard}
The Diverse People Dataset Data Card can be found in Table~\ref{tab:datacard}. The data can be found at \url{https://storage.mtls.cloud.google.com/paths_for_diversity/DPD_queries.csv}

\section{Bard Nouns and Adjectives Data Card}\label{app:datacard_bard}
The Bard Nounds and Adjectives Data Card can be found in Table~\ref{tab:datacard_bard}. The data can be downloaded as follows:
\begin{itemize}
    \item Bard prompts: \url{https://storage.mtls.cloud.google.com/paths_for_diversity/prompts.csv}
    \item Nouns and locations: \url{https://storage.mtls.cloud.google.com/paths_for_diversity/nouns_locations.csv}
    \item Adjectives: \url{https://storage.mtls.cloud.google.com/paths_for_diversity/adjectives.csv}
\end{itemize}

\begin{longtable}{p{0.25\textwidth} | p{0.3\textwidth} | p{0.35\textwidth}}
\caption{Diverse People Dataset (DPD) Datacard}
\label{tab:datacard}\\
\hline
\toprule
\dct{Publishers}\newline
Anonymous for Submission &

\dct{Team}\newline
Anonymous for Submission
\newline &

\dct{Contact Detail}\newline
\textbf{Contact}: Anonymous for Submission\newline
\textbf{Download}: Included in supplementary materials \newline \\
\midrule

\dct{Data Subject(s)}\newline
Text queries about people\newline
 & 
\dct{Data Snapshot}\newline
\begin{tabular}{lr} 
    Dataset size & 15 KB \\
    Queries & 100 \\
    Sub-queries & 498 \\
    Diversity sub-queries & 365 \\
    Irrelevant sub-queries & 133 \\
    Roles \& Professions & 33 \\
    Fashion \& Beauty & 24 \\
    Adjectives/descriptors & 18 \\
    Misc person references & 16 \\
    Events & 9
\end{tabular} &
\dct{Data Description}\newline
The dataset contains a list of queries that produce images of people. For each query there are a number of diversity-adding sub-queries -- queries in the same topic as the main query that includes additional diversity adding context. Similarly there are a number of irrelevant sub-queries which are designed to \emph{not} add additional diversity images (\eg main query: family, diversity-adding sub-query: gay family, irrelevant sub-query: family eating).\newline
\newline
These queries were used to scrape Google Image Search during May 2023. The images were not released to ensure images removed by authors are not stored and re-released. The queries were released to enable replication of our experiments.
\newline \\
\midrule

\dct{Primary Data Modality}\newline
Text Data\newline &

\dct{Link to Data}\newline
Found in supplementary material \newline &

\dct{Data Fields} \newline
Data field for each query can be found in a csv with the following headers (Header descriptions follow the format \texttt{Field Name}, \emph{Example}, Description)\newline
\begin{itemize}[leftmargin=*,noitemsep,topsep=0pt]
    \item \texttt{query type}, \emph{Roles \& professions}, Type of query.
    \item \texttt{Can every type of person fulfill this query from a visual standpoint?}, \emph{Y}, If the sub-queries are visualizable.
    \item \texttt{Query seeks multiple people}, \emph{N}, If the query specifically looks for multiple people.
    \item \texttt{query}, \emph{dancer}, The main image search query
    \item \texttt{diversity subquery 1/2/3/4}, \emph{plus size dancer}, Diversity-adding sub-query. This corresponds to 4 columns
    \item \texttt{irrelevant subquery 1/2}, \emph{dancer outside}, Irrelevant/not diversity-adding sub-query. This corresponds to 2 columns.
\end{itemize}
\\
\midrule

\dct{Purpose(s)}\newline
Replication of research \newline &

\dct{Domain(s) of application} \newline
\texttt{Research, Diversity, Evaluation} \newline &

\dct{Motivating factor(s)}
\begin{itemize}[leftmargin=*,noitemsep,topsep=0pt]
    \item Releasing the queries allows researchers to replicate this research.
    \item Not releasing the images preserves the rights of the image publishers.
\end{itemize} \\

\dct{Dataset Use(s)} \newline
Research \newline
Replication of results \newline
Diversity evaluation \newline&

\dct{Intended and suitable use case(s)}
\begin{itemize}[leftmargin=*,noitemsep,topsep=0pt]
    \item \textbf{Research}: Fairness research, Diversity research, recommendation research.
    \item \textbf{Replication}: Can be used to replicate our results.
    \item \textbf{Diversity evaluation}: Can be used to evaluate diversity of recommendation/retrieval models.
\end{itemize} &

\dct{Unsuitable use case(s)}
\begin{itemize}[leftmargin=*,noitemsep,topsep=0pt]
    \item \textbf{Publishing images}: users of this dataset should be sure to check copywrite on every image they collect by using this query dataset.
\end{itemize} \\
\midrule

\dct{Version Status}\newline
\textbf{Static Dataset} \newline
No new versions will be made available, as this represents queries scraped May 2023. \newline &

\dct{Dataset version}\newline
\begin{tabular}{ll}
    \textbf{Current Version} & 1.0 \\
    \textbf{Last Updated} & 05/2023 \\
    \textbf{Release Date} & 10/2023
\end{tabular} &

\dct{Maintenance plan}
\begin{itemize}[leftmargin=*,noitemsep,topsep=0pt]
    \item \textbf{Storage}: The dataset can be found in the supplemental materials
    \item \textbf{Versioning}: No new versions of this dataset will be made available
    \item \textbf{Availability}: The dataset will be available in the supllemental materials.
\end{itemize} \\
\bottomrule

\end{longtable}

\begin{longtable}{p{0.25\textwidth} | p{0.3\textwidth} | p{0.35\textwidth}}
\caption{Bard Nouns, Locations, and Adjectives Datacard}
\label{tab:datacard_bard}\\
\hline
\toprule
\dct{Publishers}\newline
Anonymous for Submission &

\dct{Team}\newline
Anonymous for Submission
\newline &

\dct{Contact Detail}\newline
\textbf{Contact}: Anonymous for Submission\newline
\textbf{Download}: Included in supplementary materials. Prompts can be found in the file \texttt{Prompts.csv}. Nouns and locations can be found in the file \texttt{NounsLocations.csv}. Adjectives can be found in the file \texttt{Adjectives.csv}. \newline \\
\midrule

\dct{Data Subject(s)}\newline
Text describing people & 
\dct{Data Snapshot}\newline
\newcolumntype{L}[1]{>{\raggedright\arraybackslash}p{#1}}
\begin{tabular}{L{0.2\textwidth}r} 
    Dataset size & 10 KB \\
    Prompts & 15 \\
    Adjectives & 184 \\
    Nouns & 178 \\
    Locations & 97 \\
    Age Adjectives & 8 \\
    Body Type Adjectives &	59 \\
    Disability Adjectives &	26 \\
    Ethnicity Adjectives &	11 \\
    Nationality Adjectives &	63 \\
    Religion Adjectives &	8 \\
    Sexual Orientation Adjectives &	9
\end{tabular} &
\dct{Data Description}\newline
The dataset was formed by prompting Bard to create lists of words to describe adjectives, nouns, and locations. We queried Bard with the provided prompts May 2023. We asked bard to create lists of words that satisfied a given prompt. Specifically we asked bard to produce a) a list of nouns describing types of people, b) adjectives describing diversity attributes (\eg age, body shape), and c) a list of generic locations (\eg beach, park). We removed words from the dataset that were not generic, were not appropriate, or were considered harmful.\newline\newline
Note: Bard is experimental and the results from the provided prompt may change over time. This dataset was gathered May 2023.
\newline \\
\midrule

\dct{Primary Data Modality}\newline
Text Data\newline &

\dct{Link to Data}\newline
Found in supplementary material \newline
Prompts: \texttt{Prompts.csv} \newline
Nouns+Locations: \texttt{NounsLocations.csv} \newline
Adjectives: \texttt{Adjectives.csv} \newline&

\dct{Data Fields} \newline
Data field for each query can be found in a csv with the following headers (Header descriptions follow the format \texttt{Field Name}, \emph{Example}, Description). Data fields are broken up into the different .csv files.\newline

\textbf{Prompts.csv}:
\begin{itemize}[leftmargin=*,noitemsep,topsep=0pt]
    \item \texttt{Prompt Type}, \emph{Age}, Type of prompt and the id of the conversation.
    \item \texttt{Conversation Order}, \emph{1}, The order in which prompts appeared in a conversation with Bard.
    \item \texttt{Prompt}, \emph{Can you give me a list of adjectives that describe different ages of people?}, The specific prompt text used in the conversation with Bard.
\end{itemize}
\textbf{NounsLocations.csv}
\begin{itemize}[leftmargin=*,noitemsep,topsep=0pt]
    \item \texttt{Type}, \emph{Noun}, Whether the text is a person-noun or a location.
    \item \texttt{Text}, \emph{doctor}, The text of the person-noun or location.
\end{itemize}
\textbf{Adjectives.csv}
\begin{itemize}[leftmargin=*,noitemsep,topsep=0pt]
    \item \texttt{Type}, \emph{Age}, The diversity attribute type.
    \item \texttt{Text}, \emph{infant}, The text of the diversity attribute adjective.
\end{itemize}
\\
\midrule

\dct{Purpose(s)}\newline
Replication of research \newline &

\dct{Domain(s) of application} \newline
\texttt{Research, Diversity, Evaluation} \newline &

\dct{Motivating factor(s)}
\begin{itemize}[leftmargin=*,noitemsep,topsep=0pt]
    \item Releasing the text sets and prompts allows researchers to replicate this research.
\end{itemize} \\

\dct{Dataset Use(s)} \newline
Research \newline
Replication of results \newline&

\dct{Intended and suitable use case(s)}
\begin{itemize}[leftmargin=*,noitemsep,topsep=0pt]
    \item \textbf{Research}: Fairness research, Diversity research, recommendation research.
    \item \textbf{Replication}: Can be used to replicate our results.
\end{itemize} &

\dct{Unsuitable use case(s)}
 \\
\midrule

\dct{Version Status}\newline
\textbf{Static Dataset} \newline
No new versions will be made available, as this represents conversations with bard and the resulting text sets from May 2023. \newline &

\dct{Dataset version}\newline
\begin{tabular}{ll}
    \textbf{Current Version} & 1.0 \\
    \textbf{Last Updated} & 05/2023 \\
    \textbf{Release Date} & 10/2023
\end{tabular} &

\dct{Maintenance plan}
\begin{itemize}[leftmargin=*,noitemsep,topsep=0pt]
    \item \textbf{Storage}: The dataset can be found in the supplemental materials
    \item \textbf{Versioning}: No new versions of this dataset will be made available
    \item \textbf{Availability}: The dataset will be available in the supllemental materials.
\end{itemize} \\
\bottomrule

\end{longtable}

\end{document}